\pdfoutput=1

\documentclass[11pt]{article}

\usepackage[]{ACL2023}

\usepackage{times}
\usepackage{latexsym}

\usepackage[T1]{fontenc}

\usepackage[utf8]{inputenc}

\usepackage{microtype}

\usepackage{inconsolata}

\usepackage{booktabs}
\usepackage{xcolor, colortbl}
\usepackage{pgf} 
\usepackage{tikz}
\usepackage{amsmath}
\usepackage{amssymb}
\usepackage{enumitem}
\usepackage{multicol}
\usepackage{multirow}
\usepackage{fvextra}
\usepackage{txfonts}

\title{Language Distances are Practical for Equitable Cross-Lingual Transfer}

\author{
 \textbf{York Hay Ng$^{\varheartsuit\hspace{1pt}*}$} \,
 \textbf{Razan Ahsan Rifandi$^{\varheartsuit\hspace{1pt}*}$} \,
 \textbf{Aditya Khan$^{\varheartsuit\hspace{1pt}*}$} \,
 \textbf{En-Shiun Annie Lee$^{\varheartsuit\hspace{1pt}\clubsuit}$}
\\
 $^{\varheartsuit}$University of Toronto, Canada\quad
 $^{\clubsuit}$Ontario Tech University, Canada\\
 \\
 \texttt{york.ng@mila.quebec, razan.rifandi@mail.utoronto.ca,}\\
 \texttt{aditya.khan@columbia.edu}
}

\newif\ifshowcomments
\showcommentstrue 

\begin{document}
\maketitle
\begingroup
\renewcommand\thefootnote{\fnsymbol{footnote}}
\footnotetext[1]{The authors contributed equally.}
\endgroup
\begin{abstract}
Cross-lingual transfer is strongly conditional on how the source language is chosen, but it is impractical to determine the best candidate source for every target language, especially for low-resource target languages. Language distances are widely used to rank candidate sources due to their correlation with transfer efficacy and applicability in resource-sparse settings. However, the reliability of distance-based rankers across tasks and resource levels remains underexplored. We therefore present the first equity-focused evaluation of paradigms for ranking source languages, studying resource-level inequality and task inequality across ten cross-lingual tasks and two multilingual models. While both inequalities are most pronounced for individual language distances and an English-always baseline, they are substantially reduced by training-free composite distances, and nearly eliminated by trained rankers. We further demonstrate the reliability of rankers using language distances compared to rankers using language model internals. Overall, we find that language distances provide a practical basis for equitable and performant transfer language selection. We recommend using trained rankers when task-specific transfer evaluations are available, and composite distances otherwise.
\end{abstract}

\section{Introduction}

Cross-lingual transfer supports NLP systems in languages without sufficient training data, by transferring knowledge from a higher-resource source language \citep{conneau-etal-2020-unsupervised, pires2019multilingual, wu-dredze-2019-beto}. The common default in NLP has been to transfer from English, as in standard benchmarks \citep{hu2020xtreme, liang-etal-2020-xglue}. However, this is suboptimal as transfer performance is highly dependent on the choice of source language (or transfer language) \citep{lauscher2020zero, turc2021revisiting, eronen2026language}. Language distances, often derived from typological, geographic or phylogenetic language features \citep{littell2017uriel, khan2025uriel+, goot-etal-2025-distals, eronen2026language}, are known to correlate with transfer performance \citep{philippy-etal-2023-identifying, lauscher2020zero} and are thus widely used for transfer language selection \citep{lin-etal-2019-choosing, blaschke-etal-2025-analyzing}.

However, two inequalities hamper the efficacy of distance-based selection. First, no single language distance is best across all NLP tasks, as the most predictive measure varies by task \citep{blaschke-etal-2025-analyzing, philippy-etal-2023-towards, ng-etal-2026-modality}. Second, transfer performance varies with the target language's resource level \citep{joshi-etal-2020-state, wu-dredze-2020-languages}, but whether this disparity also affects transfer language selection has not been studied systematically.

\begin{figure}
    \centering
    \includegraphics[width=\linewidth]{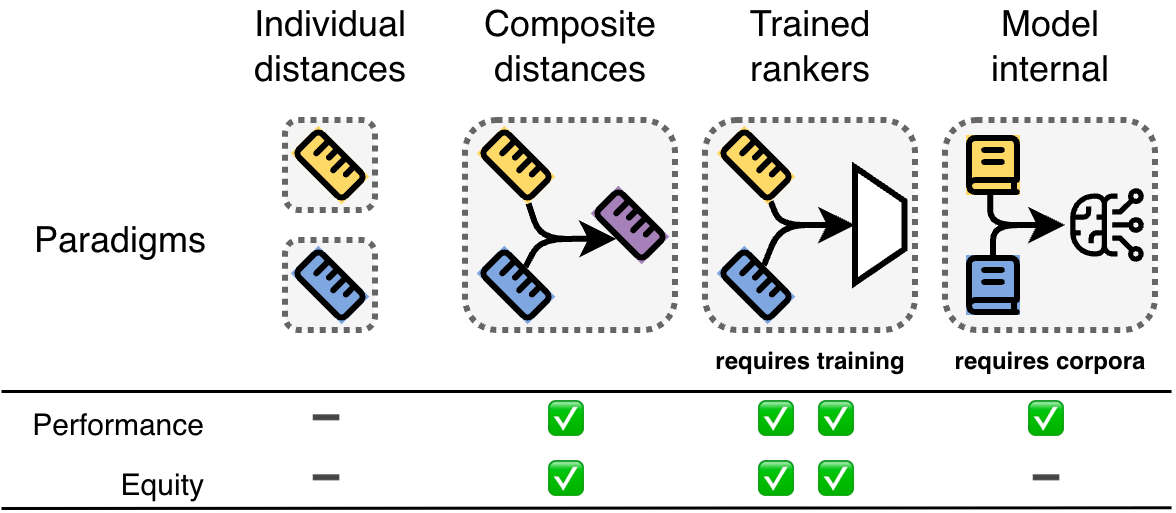}
    \caption{We evaluate four source selection paradigms on performance and task/resource-level inequality.}
    \label{fig:placeholder}
\end{figure}
There exist several paradigms for transfer language selection, each with different resource requirements. The simplest uses \textit{individual language distances} to rank source languages \citep{blaschke-etal-2025-analyzing, eronen2023zero}. \citet{ng-etal-2026-modality} proposed averaging individual distances into a single \textit{composite distance}. \textit{Trained rankers} fit a model on observed cross-lingual performance scores, but this requires task-specific model evaluation data and ranker training \citep{lin-etal-2019-choosing, rice-etal-2025-untangling}. \textit{Model internal methods} such as NNRank \citep{ebrahimi-etal-2025-model} rank source languages using the hidden representations of multilingual language models, requiring model access and parallel texts. However, these paradigms have rarely been compared under a common evaluation setup.

In this work, we perform the first equity-focused evaluation of transfer language selection based on language distances, covering 10 cross-lingual tasks and two multilingual language models, XLM-R \citep{conneau-etal-2020-unsupervised} and mT5 \citep{xue-etal-2021-mt5}, simultaneously comparing four selection paradigms under \emph{performance loss}, measuring loss relative to an oracle source language \citep{ng-etal-2026-modality}. Our contributions are:

\begin{enumerate}[itemsep=1pt]
    \item We demonstrate that the gap in performance loss between high- and low-resource target languages is mitigated through rankers using language distances, with the gap closing entirely with LightGBM on XLM-R transfer.
    \item We confirm that individual language distances exhibit task-dependence \citep{blaschke-etal-2025-analyzing}, as does transfer always from English, but show that this variance is reduced by composite distances and further reduced by trained rankers.
    \item We find that NNRank, while competitive with composite distances, retains a large performance gap on low-resource languages.
    \item We argue that language distances are practical for transfer language selection: they are dataset-independent, do not require language models nor target language corpora, and offer broad language coverage. We recommend trained rankers where transfer evaluations for the specific task are available. For unseen tasks, we recommend composite distances.
\end{enumerate}

\begin{table*}[t]
\centering
\setlength{\tabcolsep}{5pt}
\renewcommand{\arraystretch}{1.05}
\footnotesize
\begin{tabular}{llllrrrrr}
\toprule
Task & Type & Model & Metric & \#Sources & \#Targets & HRL & MRL & LRL \\
\midrule
\multicolumn{9}{l}{\textit{Sentence-level classification}} \\
SIB-200 & Topic classification & XLM-R, mT5 & F1 & 176 & 185 & 7 & 42 & 136 \\
Taxi1500 & Topic classification & XLM-R, mT5 & F1 & 32 & 762 & 8 & 28 & 726 \\
XNLI & Natural language inference & XLM-R, mT5 & F1 & 15 & 15 & 6 & 8 & 1 \\
\midrule
\multicolumn{9}{l}{\textit{Sequence labeling}} \\
UD-POS & Part-of-speech tagging & XLM-R, mT5 & F1 & 73 & 147 & 7 & 40 & 100 \\
UD-Dep & Dependency parsing & XLM-R & LAS & 73 & 147 & 7 & 40 & 100 \\
WikiANN & Named entity recognition & XLM-R, mT5 & F1 & 55 & 144 & 7 & 43 & 94 \\
MasakhaNER & Named entity recognition & XLM-R, mT5 & F1 & 19 & 19 & 0 & 0 & 19 \\
\midrule
\multicolumn{9}{l}{\textit{Question answering}} \\
TyDiQA & Question Answering & XLM-R, mT5 & F1 & 9 & 9 & 2 & 5 & 2 \\
XQuAD & Question Answering & XLM-R, mT5 & F1 & 12 & 12 & 5 & 7 & 0 \\
\midrule
\multicolumn{9}{l}{\textit{Generation}} \\
OPUS-100 & Machine Translation & mT5 & BLEU & 38 & 87 & 6 & 40 & 41 \\
\bottomrule
\end{tabular}
\caption{The tasks included in our study, and their language coverage. \# Sources shows the number of source languages. \# Targets shows the number of target languages, broken down by resource level in subsequent columns.}
\label{tab:coverage}
\end{table*}

\section{Related Work}
\subsection{Language Distances for Cross-Lingual Transfer}

Language distances are often derived from linguistic features (typological, phylogenetic, geographic), compiled from databases such as WALS \citep{wals}, Grambank \citep{grambank} and Glottolog \citep{Glottolog}. Linguistic knowledge bases such as URIEL \citep{littell2017uriel} and URIEL+ \citep{khan2025uriel+} compute distances over language vectors containing these features. \citet{ng-etal-2026-modality} proposed new distance formulations for URIEL+, while \citet{shipton-etal-2026-simple} measures differences in writing systems. Adjacent measures include ASJP \citep{asjp}, which computes lexical distances over word lists, and cognate-based approaches \citep{beaufils2025stochastic}. Language distance toolkits usually support thousands of languages while being computationally inexpensive, making them particularly valuable for low-resource languages.

Language distances have been demonstrated to correlate with cross-lingual transfer performance across tasks and language pairs \citep{lauscher2020zero, philippy-etal-2023-identifying, ahuja-etal-2022-calibration, de-vries-etal-2022-make}, and are thus widely used to choose source languages. Other applications relating to cross-lingual transfer include predicting the performance of language models \citep{anugraha2025proxylm, xia-etal-2020-predicting}, regularizing multilingual language models \citep{adilazuarda-etal-2024-lingualchemy} and aggregating language adapters \citep{chronopoulou-etal-2024-language, accou-poelman-2026-tipa}, etc.

\subsection{Transfer Language Selection}
Transfer language selection is the task of choosing source language(s) from a candidate pool to maximize cross-lingual transfer performance on a given target language. \citet{lin-etal-2019-choosing} formalized this as a ranking problem, motivated by the observation that the choice of source language substantially affects transfer performance \citep{turc2021revisiting}, but that a brute-force search is infeasible. We identify several paradigms for this task, differing in what information is used to produce a ranking.

\paragraph{Individual language distances.}

The simplest approach involves a single distance measure, where the source language with the lowest distance to the target is chosen. \citet{eronen2023zero} found a significant improvement when choosing source languages with language distances such as lang2vec \citep{littell2017uriel}, compared to transfer from English. \citet{blaschke-etal-2025-analyzing} confirmed at scale that language distance measures are predictive of cross-lingual performance. This approach is widely used, e.g. in natural language understanding tasks \citep{adelani-etal-2022-masakhaner, idris-etal-2026-comparing}, abusive language detection \citep{eronen2026language}, semantic relatedness detection \citep{zhou-etal-2024-mainlp}, and information extraction \citep{ngo2024zero}.

\paragraph{Composite distances.}
\citet{ng-etal-2026-modality} proposed averaging individual typological, geographic and genetic distances into a single composite distance, demonstrating robust performance across tasks even when compared to trained rankers.

\paragraph{Trained rankers.}
\citet{lin-etal-2019-choosing} fit gradient-boosted tree models over language distance features, demonstrating high performance in some tasks even without dataset-dependent features (e.g. word overlap). \citet{rice-etal-2025-untangling} applied this framework to part-of-speech tagging, notably finding that typological vector representations outperform distances for source selection. To that end, \citet{srinivasan2021predicting} trained XGBoost rankers on typological distances alongside WALS \citep{wals} features. These methods generally require evaluation data (of language model performance) and training ranking models.

\paragraph{Model-internal methods.}
NNRank \citep{ebrahimi-etal-2025-model} ranks source languages using multilingual language model representations, requiring model access and unlabelled corpora. \citet{chen-ritter-2021-model} similarly use model-internal representations for selecting model checkpoints. We implement NNRank as a training-free, model-internal comparison to the above distance-based methods.

\subsection{Inequalities in Cross-Lingual Transfer}
We evaluate different selection paradigms on two major inequalities.

\paragraph{Task-dependence.}
In spite of the utility of language distances, no single measure is best for all tasks. \citet{blaschke-etal-2025-analyzing} find that the most useful distance measure depends on the task and input representation. \citet{philippy-etal-2023-towards} documented contradictory findings on how linguistic similarity affects cross-lingual transfer. Moreover, \citet{dolicki2021analysing} showed that syntactic features differ in value between tasks. \citet{ploeger-etal-2025-cross} and \citet{rice-etal-2025-untangling} corroborate this task-dependence for machine translation and part-of-speech tagging, respectively.

\paragraph{Resource level inequality.}
Moreover, cross-lingual transfer is inherently challenging for low-resource languages due to the lack of data \citep{joshi-etal-2020-state, wu-dredze-2020-languages}. Indeed, \citet{malkin-etal-2022-balanced} showed that certain source languages are effective across many targets while others transfer only narrowly. Yet, the disparity this brings to the utility of transfer language selection across resource levels remains underexplored.

\section{Experimental Setup}\label{sec:exp-setup}

Following \citet{lin-etal-2019-choosing}, we evaluate transfer language selection as a ranking problem. For each task and language model, a target language $\ell$ defines a query, and each source language $s \in \mathcal{S}_{\ell}$ defines a candidate. A ranker observes features for each $(s,\ell)$ pair and ranks the candidate sources. Our evaluation asks whether high-performing transfer sources appear near the top, reflecting the operational problem: \emph{before training or evaluating many transfer systems for a new target language, which source language should be tried first?} 

\subsection{Task Setup}

Table \ref{tab:coverage} summarises the tasks, model families, evaluation metrics, and language coverage. The benchmark contains multilingual NLP tasks spanning four broad task types: \textbf{1)} sentence-level classification (SIB-200 \citep{adelani-etal-2024-sib}, Taxi1500 \citep{ma-etal-2025-taxi1500} and XNLI \citep{conneau-etal-2018-xnli}), \textbf{2)} sequence labeling (UD-POS and UD-Dep from Universal Dependencies v2.14 \citep{ud}, WikiANN \citep{pan-etal-2017-cross} and MasakhaNER \citep{adelani-etal-2022-masakhaner}), \textbf{3)} question answering (TyDiQA \citep{clark-etal-2020-tydi} and XQuAD \citep{Artetxe:etal:2019}), and \textbf{4)} generation (OPUS-100 \citep{zhang-etal-2020-improving}). We benchmark the zero-shot performance of two multilingual language models: XLM-R \citep{conneau-etal-2020-unsupervised} and mT5 \citep{xue-etal-2021-mt5}, by separately training on each source language and evaluating on all target languages.

To avoid the computational cost of evaluating on all languages for large-scale tasks, we follow \citet{ng-etal-2026-modality} for the set of source languages in Taxi1500 and SIB-200 and \citet{blaschke-etal-2025-analyzing} for both UD tasks. For WikiANN and OPUS-100, we restrict to source languages with at least 5,000 and 1,000,000 (the maximum number of samples in OPUS-100) training examples respectively, mimicking the methodology of \citet{ebrahimi-etal-2025-model} albeit with no test sample size filtering. We further filter out languages where any of the six distances could not be computed to ensure no missingness.\footnote{Filtering does not affect our resource-level findings; see Appendix \ref{app:filtering}.}

All evaluations are performed by leave-one-target-out cross-validation within each task-model setting. In each fold, one target language is held out as the query, and all candidate source languages for that target are ranked. For trained rankers, 20\% of target languages is held out as a validation split, and the remaining targets form the train split.

We evaluate rankers on performance loss \citep{ng-etal-2026-modality}. Let $p_{\ell s}$ denote the transfer performance (i.e. F1, LAS, or BLEU) from source language $s$ to target language $\ell$. Let $s^{\star}_{\ell} \in \arg\max_{s \in \mathcal{S}_{\ell}} p_{\ell s}$ be an oracle-best source, and let $\widehat{s}_{\ell}$ be the top-ranked source selected by a ranker. Performance loss is then $\mathrm{PL}_{\ell}=100 \cdot (p_{\ell s^{\star}_{\ell}}-p_{\ell \widehat{s}_{\ell}})/p_{\ell s^{\star}_{\ell}}$, where $\mathrm{PL}_{\ell}=0$ indicates an oracle-best selection, and larger values indicate greater relative loss from the best source. Normalisation by oracle performance reflects how the same performance difference is more costly where oracle performance is low, as is usually the case for low-resource targets. For comparability, we report also the unnormalised performance loss, $p_{\ell s^{\star}_{\ell}}-p_{\ell \widehat{s}_{\ell}}$, when analysing resource-level inequality in Appendix \ref{app:unnorm}.  We report mean performance loss across leave-one-target-out folds.

Performance loss serves as our primary metric since it directly measures the practical cost of source language selection, as used previously in literature \citep{blaschke-etal-2025-analyzing, ng-etal-2026-modality}. Since performance loss is primarily an indicator of top-1 source selection and not of a general shortlist of strong sources, we also report, in Appendix \ref{app:ctc}, an auxiliary diagnostic called conformal trial complexity (CTC) that evaluates a shortlist of candidate sources. This provides a statistical measure of how many source language trials are needed to include a near-best source with high probability.

\subsection{Rankers}
\label{sec:rankers}
We construct our rankers with six language distances: typological, genetic and geographic distances (4k+ languages) from \citet{ng-etal-2026-modality}, script distance (7k+ languages) from \citet{shipton-etal-2026-simple}, lexical-phonetic distance from ASJP (5k+ languages) \citep{asjp}, and a distance measuring proportional differences in Wikipedia sizes from DistaLs \citep{goot-etal-2025-distals}.

First, each distance measure is used as a standalone ranker, selecting the source language with the smallest distance to the target. Composite-Equal averages all six distances equally, following the composite distance framework of \citet{ng-etal-2026-modality}. Composite-RRF applies Reciprocal Rank Fusion \citep{cormack2009reciprocal}, which scores each candidate source by the sum of its inverse rank positions across all six distances, thus favouring consistently highly-ranked languages.

\begin{table*}[t]
\centering
\setlength{\tabcolsep}{8pt}
\renewcommand{\arraystretch}{1.05}
\footnotesize
\begin{tabular}{lccccccccc}
\toprule
Method
& \multicolumn{2}{c}{SIB-200}
& \multicolumn{2}{c}{Taxi1500}
& \multicolumn{2}{c}{UD-POS}
& \multicolumn{2}{c}{WikiANN}
& \multicolumn{1}{c}{UD-Dep} \\
\cmidrule(lr){2-3}
\cmidrule(lr){4-5}
\cmidrule(lr){6-7}
\cmidrule(lr){8-9}
\cmidrule(lr){10-10}
& XLM-R & mT5
& XLM-R & mT5
& XLM-R & mT5
& XLM-R & mT5
& XLM-R \\
\midrule
\multicolumn{10}{l}{\textit{Trained rankers}} \\
LightGBM
& \cellcolor{blue!68}\textcolor{white}{24.9}
& \cellcolor{blue!24}16.6
& \cellcolor{blue!75}\textcolor{white}{\textbf{10.0}}
& \cellcolor{blue!75}\textcolor{white}{\textbf{18.9}}
& \cellcolor{blue!53}\textcolor{white}{19.0}
& \cellcolor{blue!75}\textcolor{white}{\textbf{25.6}}
& \cellcolor{blue!65}\textcolor{white}{14.5}
& \cellcolor{blue!59}\textcolor{white}{9.7}
& \cellcolor{blue!75}\textcolor{white}{\textbf{23.8}} \\
MLP-ListNet
& \cellcolor{blue!2}34.3
& \cellcolor{blue!50}14.8
& \cellcolor{blue!73}\textcolor{white}{11.7}
& \cellcolor{blue!44}35.2
& \cellcolor{blue!62}\textcolor{white}{17.5}
& \cellcolor{blue!61}\textcolor{white}{29.7}
& \cellcolor{blue!69}\textcolor{white}{14.0}
& \cellcolor{blue!59}\textcolor{white}{9.8}
& \cellcolor{blue!62}\textcolor{white}{27.3} \\
\midrule
\multicolumn{10}{l}{\textit{Composite distances}} \\
Composite-Equal
& \cellcolor{blue!47}27.9
& \cellcolor{blue!34}15.9
& \cellcolor{blue!47}29.9
& \cellcolor{blue!45}34.7
& \cellcolor{blue!71}\textcolor{white}{15.7}
& \cellcolor{blue!59}\textcolor{white}{30.3}
& \cellcolor{blue!75}\textcolor{white}{\textbf{13.4}}
& \cellcolor{blue!61}\textcolor{white}{9.7}
& \cellcolor{blue!63}\textcolor{white}{27.1} \\
Composite-RRF
& \cellcolor{blue!39}28.9
& \cellcolor{blue!44}15.2
& \cellcolor{blue!41}33.7
& \cellcolor{blue!43}35.9
& \cellcolor{blue!75}\textcolor{white}{\textbf{15.0}}
& \cellcolor{blue!64}\textcolor{white}{28.9}
& \cellcolor{blue!52}\textcolor{white}{15.9}
& \cellcolor{blue!48}10.4
& \cellcolor{blue!61}\textcolor{white}{27.5} \\
\midrule
\multicolumn{10}{l}{\textit{Individual distances}} \\
Genetic
& \cellcolor{blue!75}\textcolor{white}{\textbf{23.9}}
& \cellcolor{blue!13}17.4
& 63.1
& 58.4
& \cellcolor{blue!36}22.2
& \cellcolor{blue!47}33.6
& \cellcolor{blue!29}18.2
& 13.1
& \cellcolor{blue!44}32.1 \\
Typological
& \cellcolor{blue!48}27.6
& \cellcolor{blue!14}17.4
& \cellcolor{blue!24}46.0
& \cellcolor{blue!8}54.3
& \cellcolor{blue!24}24.3
& \cellcolor{blue!31}38.3
& \cellcolor{blue!44}16.7
& \cellcolor{blue!75}\textcolor{white}{\textbf{8.9}}
& \cellcolor{blue!42}32.6 \\
Geographic
& \cellcolor{blue!22}31.4
& \cellcolor{blue!51}\textcolor{white}{14.7}
& \cellcolor{blue!34}39.1
& \cellcolor{blue!19}48.3
& \cellcolor{blue!10}26.9
& \cellcolor{blue!18}41.9
& \cellcolor{blue!47}16.4
& \cellcolor{blue!46}10.5
& \cellcolor{blue!36}34.3 \\
Script
& \cellcolor{blue!53}\textcolor{white}{26.9}
& \cellcolor{blue!75}\textcolor{white}{\textbf{13.0}}
& \cellcolor{blue!5}59.7
& \cellcolor{blue!16}49.9
& \cellcolor{blue!8}27.3
& \cellcolor{blue!14}43.2
& \cellcolor{blue!29}18.3
& \cellcolor{blue!19}12.0
& \cellcolor{blue!2}43.1 \\
ASJP
& \cellcolor{blue!42}28.5
& \cellcolor{blue!36}15.8
& \cellcolor{blue!11}55.0
& \cellcolor{blue!5}55.6
& \cellcolor{blue!50}19.6
& \cellcolor{blue!54}\textcolor{white}{31.8}
& \cellcolor{blue!57}\textcolor{white}{15.3}
& \cellcolor{blue!20}11.9
& \cellcolor{blue!62}\textcolor{white}{27.3} \\
Wiki-size
& 34.5
& 18.3
& \cellcolor{blue!1}62.1
& \cellcolor{blue!9}53.7
& 28.7
& 47.1
& 21.4
& \cellcolor{blue!30}11.4
& 43.7 \\
\midrule
\multicolumn{10}{l}{\textit{Baseline}} \\
English-always
& \cellcolor{blue!75}\textcolor{white}{13.0}
& 19.9
& \cellcolor{blue!26}44.9
& \cellcolor{blue!2}57.2
& \cellcolor{blue!22}24.8
& \cellcolor{blue!50}32.8
& 21.8
& \cellcolor{blue!28}11.5
& \cellcolor{blue!30}35.8 \\
\bottomrule
\end{tabular}

\vspace{2pt}

\begin{tabular}{lccccccccc}
\toprule
Method
& \multicolumn{2}{c}{MasakhaNER}
& \multicolumn{2}{c}{TyDiQA}
& \multicolumn{2}{c}{XNLI}
& \multicolumn{2}{c}{XQuAD}
& \multicolumn{1}{c}{OPUS-100} \\
\cmidrule(lr){2-3}
\cmidrule(lr){4-5}
\cmidrule(lr){6-7}
\cmidrule(lr){8-9}
\cmidrule(lr){10-10}
& XLM-R & mT5
& XLM-R & mT5
& XLM-R & mT5
& XLM-R & mT5
& mT5 \\
\midrule
\multicolumn{10}{l}{\textit{Trained rankers}} \\
LightGBM
& \cellcolor{blue!75}\textcolor{white}{\textbf{7.9}}
& \cellcolor{blue!29}10.4
& \cellcolor{blue!22}16.9
& \cellcolor{blue!62}\textcolor{white}{10.8}
& \cellcolor{blue!72}\textcolor{white}{1.4}
& \cellcolor{blue!72}\textcolor{white}{3.5}
& \cellcolor{blue!75}\textcolor{white}{\textbf{32.3}}
& \cellcolor{blue!75}\textcolor{white}{\textbf{21.1}}
& \cellcolor{blue!75}\textcolor{white}{18.6} \\
MLP-ListNet
& \cellcolor{blue!69}\textcolor{white}{9.5}
& \cellcolor{blue!73}\textcolor{white}{5.4}
& \cellcolor{blue!66}\textcolor{white}{7.8}
& \cellcolor{blue!36}15.3
& \cellcolor{blue!49}1.9
& \cellcolor{blue!3}10.8
& \cellcolor{blue!74}\textcolor{white}{32.7}
& \cellcolor{blue!46}32.6
& \cellcolor{blue!75}\textcolor{white}{\textbf{18.5}} \\
\midrule
\multicolumn{10}{l}{\textit{Composite distances}} \\
Composite-Equal
& \cellcolor{blue!29}19.6
& \cellcolor{blue!18}11.6
& \cellcolor{blue!50}11.0
& \cellcolor{blue!64}\textcolor{white}{10.5}
& \cellcolor{blue!75}\textcolor{white}{\textbf{1.3}}
& \cellcolor{blue!69}\textcolor{white}{3.9}
& \cellcolor{blue!10}55.7
& \cellcolor{blue!25}40.9
& \cellcolor{blue!63}\textcolor{white}{25.4} \\
Composite-RRF
& \cellcolor{blue!48}14.7
& \cellcolor{blue!28}10.5
& \cellcolor{blue!75}\textcolor{white}{\textbf{6.1}}
& \cellcolor{blue!75}\textcolor{white}{\textbf{8.6}}
& \cellcolor{blue!71}\textcolor{white}{1.4}
& \cellcolor{blue!68}\textcolor{white}{3.9}
& \cellcolor{blue!12}55.0
& \cellcolor{blue!29}39.1
& \cellcolor{blue!62}\textcolor{white}{26.2} \\
\midrule
\multicolumn{10}{l}{\textit{Individual distances}} \\
Genetic
& \cellcolor{blue!51}\textcolor{white}{14.0}
& \cellcolor{blue!75}\textcolor{white}{\textbf{5.1}}
& 21.3
& 21.5
& \cellcolor{blue!19}2.5
& \cellcolor{blue!75}\textcolor{white}{\textbf{3.2}}
& \cellcolor{blue!20}51.9
& \cellcolor{blue!60}\textcolor{white}{27.1}
& \cellcolor{blue!65}\textcolor{white}{24.7} \\
Typological
& \cellcolor{blue!17}22.7
& 13.7
& \cellcolor{blue!64}\textcolor{white}{8.3}
& \cellcolor{blue!69}\textcolor{white}{9.7}
& \cellcolor{blue!48}1.9
& \cellcolor{blue!48}6.0
& \cellcolor{blue!26}49.8
& \cellcolor{blue!23}41.7
& \cellcolor{blue!51}\textcolor{white}{33.1} \\
Geographic
& \cellcolor{blue!59}\textcolor{white}{11.9}
& \cellcolor{blue!59}\textcolor{white}{6.9}
& \cellcolor{blue!32}14.8
& \cellcolor{blue!30}16.4
& \cellcolor{blue!69}\textcolor{white}{1.5}
& \cellcolor{blue!71}\textcolor{white}{3.7}
& \cellcolor{blue!4}57.8
& 50.5
& \cellcolor{blue!56}\textcolor{white}{29.6} \\
Script
& \cellcolor{blue!18}22.4
& \cellcolor{blue!30}10.3
& \cellcolor{blue!18}17.5
& \cellcolor{blue!48}13.2
& 2.9
& \cellcolor{blue!17}9.3
& \cellcolor{blue!8}56.2
& \cellcolor{blue!55}\textcolor{white}{29.0}
& \cellcolor{blue!41}38.9 \\
ASJP
& \cellcolor{blue!38}17.3
& \cellcolor{blue!65}\textcolor{white}{6.2}
& \cellcolor{blue!15}18.3
& \cellcolor{blue!38}14.9
& \cellcolor{blue!4}2.8
& 11.1
& \cellcolor{blue!2}58.3
& \cellcolor{blue!1}50.2
& \cellcolor{blue!69}\textcolor{white}{22.3} \\
Wiki-size
& 26.9
& \cellcolor{blue!28}10.5
& \cellcolor{blue!23}16.6
& \cellcolor{blue!51}\textcolor{white}{12.8}
& \cellcolor{blue!65}\textcolor{white}{1.5}
& \cellcolor{blue!60}\textcolor{white}{4.8}
& 59.1
& \cellcolor{blue!4}49.0
& 63.5 \\
\midrule
\multicolumn{10}{l}{\textit{Baseline}} \\
English-always
& --
& --
& \cellcolor{blue!25}16.3
& \cellcolor{blue!44}13.9
& 3.0
& \cellcolor{blue!75}\textcolor{white}{1.5}
& \cellcolor{blue!19}52.5
& \cellcolor{blue!75}\textcolor{white}{20.9}
& -- \\
\bottomrule
\end{tabular}
\caption{Performance loss (percentage points, lower is better) for source language rankers, alongside English-always transfer, on both models across all tasks. Rankers are grouped by paradigm. Cell shading is column-wise; the per-column best ranker (excluding baseline) is shown in bold.}
\label{tab:results-pl}
\end{table*}

LightGBM, a gradient-boosted decision tree model, \citep{ke2017lightgbm} follows the ranking architecture in \citet{lin-etal-2019-choosing}. We additionally train an MLP with ListNet loss \citep{cao2007learning}, a standard listwise neural ranking approach. Both trained rankers are fit separately for each (task, model) setting, using the transfer performance scores in that setting. We note that trained rankers require task-specific model evaluations where other methods do not, and are therefore only feasible when such evaluations exist. Model-internal NNRank is discussed separately in Section \ref{sec:nnrank}. Hyperparameters are shown in Appendix \ref{app:hyperparams}.

We also study transferring from English to every target, the default in much of cross-lingual transfer. This is applicable in eight tasks with English as a source language, and we exclude folds where English is itself the target.

\section{Results}
We focus on two inequalities in ranker performance: task-dependence and resource-level inequality. As such, we compare three distance-based paradigms: individual distances, composite distances, and trained rankers, as well as the English-always baseline.

\subsection{Task inequality}
Table \ref{tab:results-pl} shows that ranker performance varies substantially across tasks for individual language distances. This confirms the central empirical concern in \citet{blaschke-etal-2025-analyzing} that language distances are informative for cross-lingual transfer, but no single distance is uniformly appropriate across tasks.

\paragraph{Individual distances are informative but task-dependent.}
There are cases where an individual distance is the most performant ranker, which is the clearest signal that language distances genuinely contain a useful signal towards source language selection. Examples include genetic distance for SIB-200 with XLM-R, script distance on SIB-200 with mT5, and typological distance on WikiANN with mT5, amongst others. Noticeably, the same table also shows that while \emph{individual} distances have (sometimes optimal) task specific signals, no individual distance is systematically superior. This task-dependence echoes \citet{blaschke-etal-2025-analyzing}. 

The intraclass coefficient ICC(1) (reported over per-fold performance loss, grouped by task, in Table \ref{tab:icc-task}) highlights this inequality. Lower values of the ICC(1) indicate that less of a ranker's performance variation is explained by the task, and hence that the method performs more uniformly across tasks \cite{liljequistIntraclassCorrelationDiscussion2019}. We see precisely the opposite of this for most individual distances, whose ICC(1) scores mostly range between $0.40$ to $0.56$, much above those of composite distances and trained rankers; the exceptions being geographic and typological distances on XLM-R.

\begin{table}[t]
\centering
\setlength{\tabcolsep}{8pt}
\renewcommand{\arraystretch}{1.05}
\footnotesize
\begin{tabular}{lcc}
\toprule
Method & XLM-R & mT5 \\
\midrule
\multicolumn{3}{l}{\textit{Trained rankers}} \\
LightGBM & \cellcolor{blue!77}\textcolor{white}{0.11} & \cellcolor{blue!80}\textcolor{white}{\textbf{0.06}} \\
MLP-ListNet & \cellcolor{blue!69}\textcolor{white}{0.16} & \cellcolor{blue!51}\textcolor{white}{0.25} \\
\midrule
\multicolumn{3}{l}{\textit{Composite distances}} \\
Composite-Equal & \cellcolor{blue!80}\textcolor{white}{\textbf{0.10}} & \cellcolor{blue!52}\textcolor{white}{0.24} \\
Composite-RRF & \cellcolor{blue!75}\textcolor{white}{0.13} & \cellcolor{blue!51}\textcolor{white}{0.25} \\
\midrule
\multicolumn{3}{l}{\textit{Individual distances}} \\
Genetic & 0.54 & \cellcolor{blue!1}0.56 \\
Typological & \cellcolor{blue!59}\textcolor{white}{0.21} & \cellcolor{blue!6}0.53 \\
Geographic & \cellcolor{blue!74}\textcolor{white}{0.13} & \cellcolor{blue!27}0.39 \\
Script & \cellcolor{blue!14}0.47 & \cellcolor{blue!11}0.49 \\
ASJP & \cellcolor{blue!27}0.40 & \cellcolor{blue!9}0.51 \\
Wiki-size & \cellcolor{blue!16}0.46 & 0.56 \\
\midrule
\multicolumn{3}{l}{\textit{Baseline}} \\
English-always & \cellcolor{blue!24}0.41 & 0.66 \\
\bottomrule
\end{tabular}
\caption{Intraclass correlation coefficient (ICC(1)) of per-fold performance loss grouped by task, for each ranking method. Lower ICC indicates more uniform performance across tasks. Cell shading is column-wise; per-column lowest is in bold.}
\label{tab:icc-task}
\end{table}

\paragraph{English-always is competitive, but as task-dependent as individual distances.}
Similar to individual distances, English-always can be the best choice for some setups (SIB-200 with XLM-R, XNLI with mT5, and XQuAD with mT5), but more often it can be among the worst performant. Moreover, performance with English-always transfer remains as task-dependent as with individual distances. English-always transfer exhibits an ICC(1) of $0.66$ on mT5, the highest of any method, and $0.41$ on XLM-R, which is within the range of individual distances. Therefore, while transfer from English may be beneficial for some tasks, a practitioner cannot know in advance whether it is a good choice for a new task.

\paragraph{Composite distances reduce task inequality without training.}
On XLM-R, Composite-Equal exhibits the lowest ICC in Table \ref{tab:icc-task}, $0.10$, and Composite-RRF is close at $0.13$. Across both models, the ICC values of composite distances are substantially lower than the ICC values of individual distances and English-always transfer, and are comparable to the trained rankers.

Crucially, the absolute results in Table \ref{tab:results-pl} show that this reduction in task inequality also comes with competitive performance to trained rankers. On UD-POS with XLM-R, Composite-RRF obtains the lowest performance loss, and composite distances significantly outperform trained rankers.\footnote{See Appendix \ref{app:significance}.} This is similar to WikiANN with XLM-R, where Composite-Equal obtains the lowest performance loss. In the smaller-coverage tasks, Composite-RRF is strongest on TyDiQA, with performance loss $6.1$ for XLM-R and $8.6$ for mT5.

Our results therefore extend the individual-distance conclusions of \citet{blaschke-etal-2025-analyzing}. If the practitioner must choose one individual distance measure, task-dependence poses a serious obstacle to performance stability. However, by simply combining multiple distance measures, much of this task inequality is reduced without requiring training ranker models for each task.

\paragraph{Trained rankers further reduce task inequality.}
Trained rankers provide a second way to reduce task inequality. They provide uniformly low ICC(1) values in Table \ref{tab:icc-task}, notably with LightGBM exhibiting an ICC of only $0.06$ on mT5, substantially below any other ranker. Simultaneously, trained rankers attain the best performance on 8 of the 18 columns in Table \ref{tab:results-pl}, more than any other paradigm. We also find that LightGBM performs best on smaller-coverage tasks in Table \ref{tab:results-pl}, including MasakhaNER with XLM-R (PL $7.9$), XQuAD with XLM-R (PL $32.3$), and XQuAD with mT5 (PL $21.1$). 

\begin{figure*}[t]
    \centering
    \includegraphics[width=\textwidth]{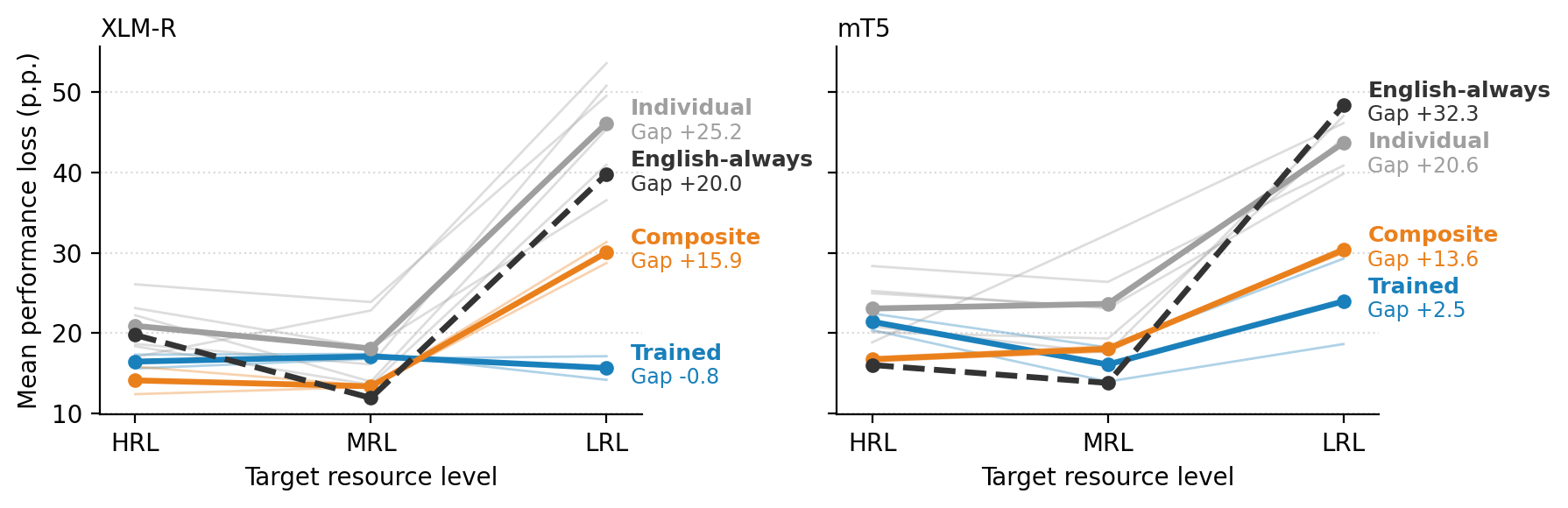}
    \caption{Mean per-fold performance loss (percentage points, lower is better) by the target language's resource level (HRL/MRL/LRL), pooled across tasks. Bold lines show the group mean; lines underneath show the trajectory of individual rankers. \textbf{Gap} is defined as the LRL minus HRL mean performance loss for that group, lower is better.}
    \label{fig:resource-inequality}
\end{figure*}

Altogether, this presents a hierarchy of inequality: 1) English-always transfer and individual distances are sometimes optimal but performance is task-dependent, 2) composite distances reduce task dependency by simply averaging distances, and 3) trained rankers reduce it even more with supervision, allowing the model to learn complex relationships between these distances. This hierarchy is largely reflected in overall performance as well.\footnote{We verify this with significance testing in Appendix \ref{app:significance}.}

\subsection{Resource level inequality}
\label{sec:resource}
We next consider whether ranker performance varies with the target language's resource level. Figure \ref{fig:resource-inequality} summarises the resource-level pattern by plotting mean per-fold performance loss across HRL, MRL, and LRL targets.\footnote{We take \citet{khan2025uriel+}'s definition, using \citet{joshi-etal-2020-state}'s classifications: languages in classes 0-2 are LRLs, 3-4 are MRLs, and languages in class 5 are HRLs.} A smaller LRL-HRL gap implies a more equitable ranker. We further report the corresponding analysis with unnormalised loss in Appendix \ref{app:unnorm}.

\paragraph{Individual distances and English-always are least equitable.}
Individual distances and English-always transfer exhibit the largest resource-level inequality. In Figure \ref{fig:resource-inequality}, the mean curve for individual distances rises sharply from HRL to LRL targets for both model families, with gaps ranging from +12.5 (script, mT5) to +36.6 (Wiki-size, XLM-R). This gap arises due to individual distances incurring a similarly high absolute performance difference from the oracle on both HRL and LRL targets, while oracle performance is substantially lower on the latter, as shown in Appendix \ref{app:unnorm}. English-always shows a similar pattern, with gaps of +20.0 on XLM-R and +32.3 on mT5; the latter gap exceeds that of any individual distance on mT5, despite English-always attaining the lowest loss in MRL of all paradigms on both models. This showcases their limited reliability for low-resource targets where the practical need for transfer language selection is greatest.\footnote{Appendix \ref{app:lrl-selection} demonstrates that while oracle performance is lower for LRL targets, selecting a good source language remains important.}

This result augments the task-dependence finding. Even if an individual distance is useful for a task, it may still produce unequal losses for low-resource targets. Therefore, selection methods should be evaluated not only by absolute performance or by task-level variation, but also by the distribution of losses across resource levels.

\paragraph{Training-free composite rankers reduce the resource gap.}
Training-free composite rankers substantially reduce the low-resource gap. In Figure \ref{fig:resource-inequality}, the mean LRL-HRL gap for the training-free group is $15.9$ on XLM-R and $13.6$ on mT5. While these gaps are still positive, they are much smaller than the corresponding gaps for individual distances and English-always transfer. This shows that averaging individual distances not only results in more consistent performance across tasks, but also results in better equity across resource levels. Since composite distances require no model access and no task-specific ranker training, they provide a computationally simple way to make transfer language selection more stable.

\paragraph{Trained rankers nearly close the resource-level gap.}
Trained rankers give the strongest reduction in resource-level inequality. In Figure \ref{fig:resource-inequality}, for LightGBM, the gap is essentially closed, with a gap of -3.1 on XLM-R and -1.7 on mT5. For MLP, a small positive gap remains (+1.5 on XLM-R and +6.8 on mT5), but it is much smaller than the gaps for English-always transfer, individual distances and training-free composite rankers. This resource-level result explains why the strong absolute performance of LightGBM in Table \ref{tab:results-pl} is practically important. They are not just improvements in average ranker quality; they also correspond to a reduction in the impact that resource level has. 

\paragraph{Inequalities are symmetric, and the solution is hierarchical.}
A key finding is that across individual task/model setups, there is often at least one distance that is highly performant in transfer language selection, but this is usually also paired with a strong preference for higher-resource targets. Once these individual strong distances are averaged together, one can simultaneously reduce the resource bias, while taking steps towards equalising performance across model/task setups. This can be done without training ranking models. Indeed, once one introduces trained rankers, the resource gap is nearly eliminated, and performance is generally strongest across models and tasks. This is a consistent hierarchy across both levels of analysis: task and resource-level.

\section{Comparison with NNRank}\label{sec:nnrank}
The rise of language models has motivated the usage of model representations across multilingual NLP, and more specifically, in transfer language selection. \citet{ebrahimi-etal-2025-model} introduced NNRank, which scores the similarity in model representations of parallel text. A key prerequisite to using NNRank is access to both multilingual model representations and parallel text, in contrast to language-distance-based rankers. Therefore, the language-distance-based rankers studied in Section \ref{sec:exp-setup} are \emph{a priori} preferable in low-resource settings. 

A natural question is how comparable language-distance-based rankers are in settings where model access and parallel texts \textit{are} available.

\subsection{Setup}
We implement NNRank \citep{ebrahimi-etal-2025-model} as a task-independent ranker akin to composite distances, sampling 100 parallel verses from the eBible corpus \citep{ebible}. For each target language, we score source languages by the similarity between parallel Bible verses in mBERT's representations at layer 8. As not all languages in our evaluation have a corresponding Bible translation, we restrict each task to only the languages with available texts (see Appendix \ref{app:nnrank-langs}). We compare against Composite-Equal and LightGBM as representatives from their respective paradigms, as they exhibited better performance and equalities overall.

\begin{table}[t]
\centering
\setlength{\tabcolsep}{5pt}
\renewcommand{\arraystretch}{1.05}
\footnotesize
\begin{tabular}{llccc}
\toprule
Task & Model & NNRank & Comp-Eq & LGBM \\
\midrule
\multirow{2}{*}{SIB-200} & XLM-R & 30.5 & \cellcolor{blue!10}28.4 & \cellcolor{blue!80}\textcolor{white}{\textbf{14.7}} \\
 & mT5 & \cellcolor{blue!4}13.7 & 13.9 & \cellcolor{blue!80}\textcolor{white}{\textbf{10.0}} \\
\midrule
\multirow{2}{*}{Taxi1500} & XLM-R & 35.9 & \cellcolor{blue!20}29.5 & \cellcolor{blue!80}\textcolor{white}{\textbf{9.8}} \\
 & mT5 & 46.4 & \cellcolor{blue!37}33.5 & \cellcolor{blue!80}\textcolor{white}{\textbf{18.6}} \\
\midrule
\multirow{2}{*}{UD-POS} & XLM-R & \cellcolor{blue!80}\textcolor{white}{\textbf{14.3}} & 21.2 & \cellcolor{blue!36}18.1 \\
 & mT5 & \cellcolor{blue!18}27.5 & 28.9 & \cellcolor{blue!80}\textcolor{white}{\textbf{22.7}} \\
\midrule
\multirow{2}{*}{WikiANN} & XLM-R & \cellcolor{blue!80}\textcolor{white}{\textbf{12.3}} & \cellcolor{blue!43}14.1 & 16.1 \\
 & mT5 & \cellcolor{blue!80}\textcolor{white}{\textbf{7.1}} & 9.5 & \cellcolor{blue!56}\textcolor{white}{7.8} \\
\midrule
UD-Dep & XLM-R & \cellcolor{blue!56}\textcolor{white}{24.9} & 26.5 & \cellcolor{blue!80}\textcolor{white}{\textbf{24.3}} \\
\midrule
\multirow{2}{*}{MasakhaNER} & XLM-R & 12.8 & \cellcolor{blue!64}\textcolor{white}{8.3} & \cellcolor{blue!80}\textcolor{white}{\textbf{7.2}} \\
 & mT5 & 5.3 & \cellcolor{blue!27}4.7 & \cellcolor{blue!80}\textcolor{white}{\textbf{3.4}} \\
\midrule
\multirow{2}{*}{TyDiQA} & XLM-R & \cellcolor{blue!80}\textcolor{white}{\textbf{8.6}} & \cellcolor{blue!45}12.3 & 17.3 \\
 & mT5 & 11.6 & \cellcolor{blue!40}10.6 & \cellcolor{blue!80}\textcolor{white}{\textbf{9.5}} \\
\midrule
\multirow{2}{*}{XNLI} & XLM-R & 1.6 & \cellcolor{blue!61}\textcolor{white}{1.4} & \cellcolor{blue!80}\textcolor{white}{\textbf{1.3}} \\
 & mT5 & \cellcolor{blue!80}\textcolor{white}{\textbf{3.7}} & 4.7 & \cellcolor{blue!39}4.2 \\
\midrule
\multirow{2}{*}{XQuAD} & XLM-R & \cellcolor{blue!30}10.2 & 14.1 & \cellcolor{blue!80}\textcolor{white}{\textbf{3.7}} \\
 & mT5 & 43.8 & \cellcolor{blue!37}31.1 & \cellcolor{blue!80}\textcolor{white}{\textbf{16.5}} \\
\midrule
OPUS-100 & mT5 & \cellcolor{blue!80}\textcolor{white}{\textbf{15.0}} & 22.4 & \cellcolor{blue!36}19.1 \\
\bottomrule
\end{tabular}
\caption{Performance loss (p.p., lower is better) for NNRank, Composite-Equal, and LightGBM on the subset of languages covered by eBible (Appendix \ref{app:nnrank-langs}). Colour shading is row-wise; per-row best is in bold.} 
\label{tab:nnrank-overall}
\end{table}

\begin{figure*}[t]
\centering
\includegraphics[width=\textwidth]{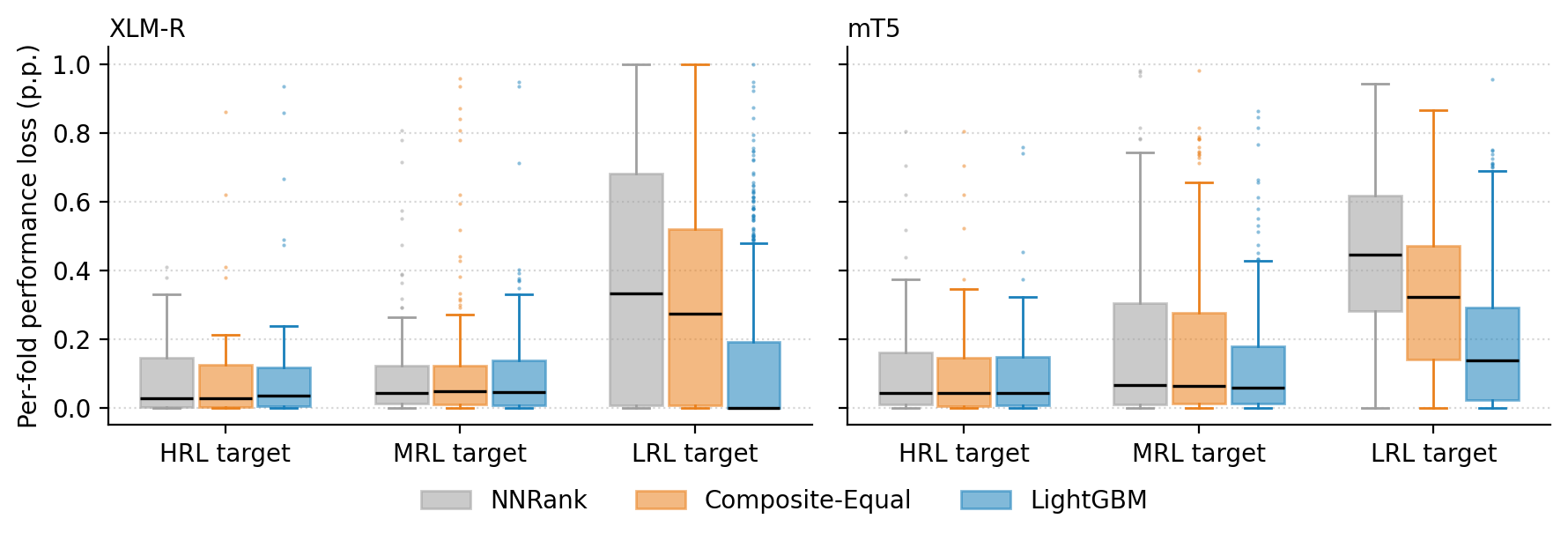}
\caption{Box plots of task-pooled performance loss (p.p., lower is better), split by target language
resource level.}
\label{fig:spread}
\end{figure*}

\subsection{Results}
Table \ref{tab:nnrank-overall} compares the performance loss of NNRank with Composite-Equal and LightGBM, while Figure \ref{fig:spread} compares their resource-level equity.

\paragraph{NNRank is competitive on several task/model pairs.}
Table \ref{tab:nnrank-overall} highlights the strength of NNRank in some settings. It exhibits lowest performance loss on at least one model for UD-POS, WikiANN, TyDiQA, XNLI, and OPUS-100. These results are consistent with the motivation of model-internal methods, as hidden representations can contain useful information about cross-lingual compatibility that is external to language distances. 

\paragraph{Distance-based methods often match or outperform NNRank.}
Simultaneously, we observe that LightGBM remains the most consistent ranker across task/model setups, achieving superior performance in at least one model for all tasks except WikiANN and OPUS-100. Likewise, Composite-Equal outperforms NNRank on almost half of task/model setups, showing that both methods are competitive.

These comparisons are important because distance-based rankers require substantially weaker inputs. Composite distances do not require model internals, task-specific model evaluation data, ranker training, or target language corpora. Trained rankers require model evaluations for related targets, but still do not require access to hidden states or model-specific representations. Thus, when LightGBM or even Composite-Equal matches or improves over NNRank, the result strengthens the practical case for language distances.

\paragraph{NNRank falls behind on resource-level equity.} 
Figure \ref{fig:spread} shows that the most prominent difference between NNRank and distance-based methods appears across resource levels. For HRL and MRL targets, all three methods have low median performance loss in both XLM-R and mT5. However, transfer language selection becomes much harder on LRL targets. In XLM-R, NNRank has a substantially higher median loss on LRL targets, while Composite-Equal is lower and LightGBM has the lowest median and tightest interquartile range. The same pattern appears for mT5, although the LRL losses are higher overall: NNRank has the highest LRL median, Composite-Equal improves on NNRank, and LightGBM improves further.

This provides some nuance to the aforementioned performance analysis between NNRank and distance-based rankers. NNRank is not weak on average; indeed, Table \ref{tab:nnrank-overall} shows that it is superior in several task-model setups. Rather, the strength of distance-based rankers lies in low-resource target languages, where transfer language selection matters most. In comparison to NNRank, distance-based rankers reduce both the median loss in LRL targets and the spread of LRL outcomes.

\paragraph{In resource-sparse settings, distance-based rankers dominate.}
The central story that the above analysis showcases is that model internal methods are most applicable for transfer language selection in higher-resourced settings. That is also true in terms of the inputs required for the ranking models. In low-resource settings, however, methods based on language distances provide performance that is at least competitive, and usually superior.

\section{Conclusion}

We perform a large-scale evaluation of four transfer language selection paradigms across ten tasks and two multilingual models. Our results reveal a consistent hierarchy across two inequalities. First, we confirm the task-dependent utility of individual language distances, similar to transferring always from English. Composite distances substantially reduce this variance without requiring training. Trained rankers reduce it further, however they are only feasible when transfer evaluations for the task are available. Second, composite distances substantially narrow the performance gap between resource levels, and trained rankers effectively close it. Model-internal selection is competitive with composite distances, but language-distance-based methods excel in low-resource languages. These findings support the practicality of language distances in transfer language selection, which require no model access nor target language corpora, yet individually cover thousands of languages and support equitable transfer.

\section*{Limitations}


\paragraph{Task assumptions.} Within transfer language selection, we restrict our evaluation to zero-shot transfer from a single source language. We acknowledge that transfer from multiple sources can yield performance improvements \citep{chen-etal-2019-multi-source, lim-etal-2024-analysis}, and represents a related selection problem over language combinations. Few-shot transfer, by fine-tuning on a small number of examples in the target language, can also narrow the zero-shot transfer gap \citep{schmidt-etal-2022-dont}.

Nevertheless, we consider zero-shot, single-source transfer a meaningful task. It is one of the setups under which transfer language selection has been most systematically studied \citep{lin-etal-2019-choosing, eronen2023zero, blaschke-etal-2025-analyzing, ng-etal-2026-modality}. We therefore leave extensions to multi-source and few-shot setups to future work.

\paragraph{Language coverage.} The distance measures we use support several thousand languages. In contrast, our evaluation is constrained by the languages supported by existing NLP datasets, which represents only a small subset of the languages supported. While we break down our analysis by resource level and find that language distances are robust for low-resource languages, there exists an inherent bias towards languages with more resources. Moreover, as the composite distance for a particular language pair requires all distances to be available for that language pair, this further restricts language coverage. Appendix \ref{app:filtering} shows that our composite distance can be unavailable for up to 16\% of languages in a particular task.

\paragraph{Model coverage.} We evaluate only on two open-source models, XLM-R and mT5. These models are demonstrated to have strong multilingual performance, and our findings hold true in cross-model comparisons. However, we leave comparisons to more recent, especially decoder-only multilingual models, to future work.

\section*{Ethical Considerations}
One central motivation of our work is to find effective rankers that are capable of overcoming issues with resource-level inequalities. Ideally, it reduces the cost of finding useful transfer sources. However, source language rankings should not be interpreted as statements about the intrinsic value or importance of languages or language communities. They are recommendations purely for the purpose of downstream tasks, and not linguistic or sociocultural judgments. 

We note that there is no new human-subject data collected or presented in this analysis. Furthermore we only make use of public datasets and models under the appropriate license they are released with. Therefore, our work may inherit implicit linguistic biases underlying these resources.

\bibliography{custom}
\bibliographystyle{acl_natbib}

\appendix

\newcommand{\nocpass}[1]{\cellcolor{green!12}#1}
\newcommand{\nocfail}[1]{\cellcolor{red!12}#1}

\begin{table*}[t]
\centering
\scriptsize
\setlength{\tabcolsep}{2.8pt}
\renewcommand{\arraystretch}{1.05}

\resizebox{\textwidth}{!}{%
\begin{tabular}{l*{12}{c}}
\toprule
Method
& \multicolumn{4}{c}{SIB-200}
& \multicolumn{4}{c}{Taxi1500}
& \multicolumn{4}{c}{UD-POS} \\
\cmidrule(lr){2-5}
\cmidrule(lr){6-9}
\cmidrule(lr){10-13}
& \multicolumn{2}{c}{XLM-R}
& \multicolumn{2}{c}{mT5}
& \multicolumn{2}{c}{XLM-R}
& \multicolumn{2}{c}{mT5}
& \multicolumn{2}{c}{XLM-R}
& \multicolumn{2}{c}{mT5} \\
\cmidrule(lr){2-3}
\cmidrule(lr){4-5}
\cmidrule(lr){6-7}
\cmidrule(lr){8-9}
\cmidrule(lr){10-11}
\cmidrule(lr){12-13}
& CTC & HR
& CTC & HR
& CTC & HR
& CTC & HR
& CTC & HR
& CTC & HR \\
\midrule
\multicolumn{13}{l}{\textit{Trained rankers}} \\
LightGBM
& \cellcolor{blue!67}\textcolor{white}{43} & \nocpass{$94.6 \pm 1.7$}
& \cellcolor{blue!73}\textcolor{white}{30} & \nocpass{$93.0 \pm 1.9$}
& \cellcolor{blue!75}\textcolor{white}{\textbf{7}} & \nocpass{$90.3 \pm 1.1$}
& \cellcolor{blue!75}\textcolor{white}{\textbf{11}} & \nocpass{$91.5 \pm 1.0$}
& \cellcolor{blue!75}\textcolor{white}{\textbf{12}} & \nocfail{$81.0 \pm 3.2$}
& \cellcolor{blue!50}45 & \nocpass{$94.6 \pm 1.9$} \\
MLP
& \cellcolor{blue!75}\textcolor{white}{\textbf{31}} & \nocpass{$93.0 \pm 1.9$}
& \cellcolor{blue!66}\textcolor{white}{34} & \nocpass{$92.4 \pm 1.9$}
& \cellcolor{blue!38}19 & \nocpass{$92.8 \pm 0.9$}
& \cellcolor{blue!69}\textcolor{white}{12} & \nocpass{$89.2 \pm 1.1$}
& \cellcolor{blue!67}\textcolor{white}{17} & \nocfail{$85.0 \pm 2.9$}
& \cellcolor{blue!70}\textcolor{white}{39} & \nocpass{$91.2 \pm 2.3$} \\
\midrule
\multicolumn{13}{l}{\textit{Composite distances}} \\
Comp-Eq
& \cellcolor{blue!16}124 & \nocpass{$96.2 \pm 1.4$}
& \cellcolor{blue!44}47 & \nocpass{$93.5 \pm 1.8$}
& \cellcolor{blue!32}21 & \nocpass{$90.4 \pm 1.1$}
& \cellcolor{blue!27}23 & \nocpass{$92.3 \pm 1.0$}
& \cellcolor{blue!66}\textcolor{white}{18} & \nocfail{$85.7 \pm 2.9$}
& \cellcolor{blue!63}\textcolor{white}{40} & \nocpass{$91.2 \pm 2.3$} \\
Comp-RRF
& \cellcolor{blue!11}131 & \nocpass{$96.2 \pm 1.4$}
& \cellcolor{blue!37}51 & \nocpass{$94.1 \pm 1.7$}
& \cellcolor{blue!24}23 & \nocpass{$90.7 \pm 1.1$}
& \cellcolor{blue!10}28 & \nocpass{$93.2 \pm 0.9$}
& \cellcolor{blue!63}\textcolor{white}{19} & \nocpass{$89.1 \pm 2.6$}
& \cellcolor{blue!75}\textcolor{white}{\textbf{37}} & \nocpass{$91.2 \pm 2.3$} \\
\midrule
\multicolumn{13}{l}{\textit{Individual distances}} \\
Genetic
& \cellcolor{blue!6}139 & \nocpass{$93.5 \pm 1.8$}
& 72 & \nocpass{$94.1 \pm 1.7$}
& \cellcolor{blue!25}23 & \nocpass{$92.0 \pm 1.0$}
& \cellcolor{blue!26}24 & \nocpass{$93.7 \pm 0.9$}
& \cellcolor{blue!57}\textcolor{white}{23} & \nocfail{$80.3 \pm 3.3$}
& \cellcolor{blue!68}\textcolor{white}{39} & \nocpass{$89.1 \pm 2.6$} \\
Typological
& \cellcolor{blue!44}80 & \nocpass{$93.0 \pm 1.9$}
& \cellcolor{blue!54}\textcolor{white}{41} & \nocpass{$93.5 \pm 1.8$}
& \cellcolor{blue!32}21 & \nocpass{$90.0 \pm 1.1$}
& \cellcolor{blue!10}28 & \nocpass{$92.5 \pm 1.0$}
& \cellcolor{blue!24}44 & \nocpass{$91.2 \pm 2.3$}
& \cellcolor{blue!50}44 & \nocpass{$91.8 \pm 2.3$} \\
Geographic
& 149 & \nocpass{$93.0 \pm 1.9$}
& \cellcolor{blue!52}42 & \nocpass{$91.9 \pm 2.0$}
& 31 & \nocpass{$95.5 \pm 0.7$}
& \cellcolor{blue!5}29 & \nocpass{$93.8 \pm 0.9$}
& \cellcolor{blue!47}30 & \nocpass{$89.8 \pm 2.5$}
& \cellcolor{blue!31}50 & \nocpass{$95.2 \pm 1.8$} \\
Script
& \cellcolor{blue!44}79 & \nocpass{$96.2 \pm 1.4$}
& \cellcolor{blue!75}\textcolor{white}{\textbf{29}} & \nocpass{$88.1 \pm 2.4$}
& \cellcolor{blue!24}23 & \nocpass{$95.9 \pm 0.7$}
& \cellcolor{blue!57}\textcolor{white}{15} & \nocpass{$90.9 \pm 1.0$}
& 60 & \nocpass{$95.2 \pm 1.8$}
& 60 & \nocpass{$93.2 \pm 2.1$} \\
ASJP
& \cellcolor{blue!31}100 & \nocpass{$95.1 \pm 1.6$}
& \cellcolor{blue!40}49 & \nocpass{$93.0 \pm 1.9$}
& \cellcolor{blue!6}29 & \nocpass{$90.8 \pm 1.0$}
& \cellcolor{blue!4}29 & \nocpass{$93.4 \pm 0.9$}
& \cellcolor{blue!49}28 & \nocfail{$84.4 \pm 3.0$}
& \cellcolor{blue!15}55 & \nocpass{$94.6 \pm 1.9$} \\
Wiki-size
& \cellcolor{blue!26}108 & \nocpass{$96.8 \pm 1.3$}
& \cellcolor{blue!56}\textcolor{white}{40} & \nocpass{$89.7 \pm 2.2$}
& \cellcolor{blue!19}25 & \nocpass{$95.7 \pm 0.7$}
& 30 & \nocpass{$90.8 \pm 1.0$}
& \cellcolor{blue!5}56 & \nocpass{$93.2 \pm 2.1$}
& \cellcolor{blue!9}57 & \nocpass{$93.2 \pm 2.1$} \\
\bottomrule
\end{tabular}%
}

\vspace{2pt}

\makebox[\textwidth][c]{%
\begin{tabular}{l*{6}{c}}
\toprule
Method
& \multicolumn{4}{c}{WikiANN}
& \multicolumn{2}{c}{UD-Dep} \\
\cmidrule(lr){2-5}
\cmidrule(lr){6-7}
& \multicolumn{2}{c}{XLM-R}
& \multicolumn{2}{c}{mT5}
& \multicolumn{2}{c}{XLM-R} \\
\cmidrule(lr){2-3}
\cmidrule(lr){4-5}
\cmidrule(lr){6-7}
& CTC & HR
& CTC & HR
& CTC & HR \\
\midrule
\multicolumn{7}{l}{\textit{Trained rankers}} \\
LightGBM
& \cellcolor{blue!67}\textcolor{white}{22} & \nocpass{$89.6 \pm 2.5$}
& \cellcolor{blue!60}\textcolor{white}{18} & \nocpass{$93.8 \pm 2.0$}
& \cellcolor{blue!71}\textcolor{white}{19} & \nocfail{$79.6 \pm 3.3$} \\
MLP
& \cellcolor{blue!56}\textcolor{white}{25} & \nocpass{$90.3 \pm 2.5$}
& \cellcolor{blue!70}\textcolor{white}{15} & \nocpass{$95.1 \pm 1.8$}
& \cellcolor{blue!75}\textcolor{white}{\textbf{17}} & \nocfail{$81.0 \pm 3.2$} \\
\midrule
\multicolumn{7}{l}{\textit{Composite distances}} \\
Comp-Eq
& \cellcolor{blue!50}26 & \nocpass{$91.0 \pm 2.4$}
& \cellcolor{blue!29}27 & \nocpass{$96.5 \pm 1.5$}
& \cellcolor{blue!55}\textcolor{white}{26} & \nocfail{$84.4 \pm 3.0$} \\
Comp-RRF
& \cellcolor{blue!47}27 & \nocpass{$88.9 \pm 2.6$}
& \cellcolor{blue!31}26 & \nocpass{$95.1 \pm 1.8$}
& \cellcolor{blue!70}\textcolor{white}{19} & \nocfail{$78.2 \pm 3.4$} \\
\midrule
\multicolumn{7}{l}{\textit{Individual distances}} \\
Genetic
& \cellcolor{blue!20}33 & \nocpass{$91.7 \pm 2.3$}
& \cellcolor{blue!75}\textcolor{white}{\textbf{14}} & \nocpass{$90.3 \pm 2.5$}
& \cellcolor{blue!54}\textcolor{white}{27} & \nocfail{$81.6 \pm 3.2$} \\
Typological
& 37 & \nocpass{$93.1 \pm 2.1$}
& \cellcolor{blue!47}22 & \nocpass{$95.8 \pm 1.7$}
& \cellcolor{blue!50}28 & \nocfail{$82.3 \pm 3.1$} \\
Geographic
& \cellcolor{blue!75}\textcolor{white}{\textbf{21}} & \nocfail{$86.8 \pm 2.8$}
& \cellcolor{blue!54}\textcolor{white}{20} & \nocpass{$94.4 \pm 1.9$}
& \cellcolor{blue!49}29 & \nocfail{$82.3 \pm 3.1$} \\
Script
& \cellcolor{blue!56}\textcolor{white}{25} & \nocfail{$85.4 \pm 2.9$}
& \cellcolor{blue!26}27 & \nocpass{$94.4 \pm 1.9$}
& \cellcolor{blue!2}50 & \nocpass{$88.4 \pm 2.6$} \\
ASJP
& \cellcolor{blue!64}\textcolor{white}{23} & \nocpass{$88.9 \pm 2.6$}
& \cellcolor{blue!8}32 & \nocpass{$95.8 \pm 1.7$}
& \cellcolor{blue!36}35 & \nocfail{$84.4 \pm 3.0$} \\
Wiki-size
& \cellcolor{blue!14}34 & \nocpass{$87.5 \pm 2.8$}
& 34 & \nocpass{$92.4 \pm 2.2$}
& 51 & \nocpass{$89.1 \pm 2.6$} \\
\bottomrule
\end{tabular}%
}

\caption{CTC is the metric; lower is better. HR is the hit-rate, the empirical percentage of held-out target language folds for which the calibrated shortlist contains a source within $5\%$ relative performance loss of the oracle-best source, reported as HR $\pm$ standard error. CTC cell shading follows the column-wise scale used in Table \ref{tab:results-pl}; HR cells are shaded green when $\mathrm{HR}+\mathrm{SE}\geq 90$ and red otherwise.}
\label{tab:appendix-cnotc-diagnostics}
\end{table*}

\section{Conformal Trial Complexity for Calibrated Source Language Shortlists}
\label{app:ctc}

Performance loss is our primary metric because it measures the cost of following a ranker's first recommendation. This is the relevant evaluation when a practitioner can try only one source language for a new target language. However, source language selection is often a shortlist problem: a practitioner may be able to try a small number of source languages, but not the full candidate pool. In that setting, top-1 performance does not fully describe the usefulness of a ranking. We also want to know how many source language trials are needed before the shortlist is likely to contain a near-best source.

We therefore define a metric in \textbf{conformal trial complexity} (CTC) as a diagnostic for calibrated source language shortlists. CTC asks: \emph{how many source languages must be evaluated to include an operationally near-best source with high probability?} 

\paragraph{Definition.}
For target language $\ell$, let $r_{\ell s}$ be the rank assigned to source $s$. Define the $\epsilon$-near-best set by $\mathcal{G}_{\ell}(\epsilon)=\{s \in \mathcal{S}_{\ell}: (p_{\ell s^\star_\ell}-p_{\ell s})/p_{\ell s^\star_\ell} \leq \epsilon\}$ and let $A_{\ell}=\min_{s \in \mathcal{G}_{\ell}(\epsilon)} r_{\ell s}$, the rank of the first near-best source. We set $\widehat{k}_{\alpha}$ via split-conformal calibration \citep{vovk2005algorithmic}, yielding a shortlist size calibrated to contain an $\epsilon$-near-best source with high probability. We choose $\alpha=0.1$ and $\epsilon=0.05$, corresponding to a nominal $90\%$ target for including a source language within $5\%$ relative performance loss of the oracle best, provided that statistical assumptions (discussed subsequently) hold. Provided that they do not, CTC remains interpretable as an adaptively chosen top-$k$ cutoff for finding an $\epsilon$-near-best source. This avoids a problem that many information retrieval metrics have, such as top-$k$ accuracy and precision@$k$, which require fixing the trial budget in advance, and hence yield different conclusions depending on the choice of $k$. 

\paragraph{Guarantee and assumptions.}
Let $A_{\ell_{\mathrm{test}}}$ denote the unobserved nonconformity score for a new test target, i.e., the rank of the first $\epsilon$-near-best source. If the calibration targets and test target are exchangeable conditional on the fitting set, the split-conformal argument gives $\mathbb{P}(A_{\ell_{\mathrm{test}}} \leq \widehat{k}_{\alpha}) \geq 1-\alpha$ \citep{vovk2005algorithmic,shafer2008tutorial,angelopoulos2023conformal}. Equivalently, the top-$\widehat{k}_{\alpha}$ shortlist contains at least one $\epsilon$-near-best source with probability at least $1-\alpha$. This guarantee is marginal over target language queries: CTC controls the size of a shortlist likely to contain an operationally acceptable source, rather than guaranteeing success for every target.

The main assumption is exchangeability of target language queries within a task-model setting. Here, a query is a target language together with its candidate source set and observed source-transfer performances. This is a fairly strong assumption, which we study in Table \ref{tab:appendix-cnotc-diagnostics}.

The ranking rule must also be fixed before evaluating the held-out target, as in split conformal prediction. For trained rankers, this is enforced by fitting only on the fitting targets and using a separate calibration set to choose $\widehat{k}_{\alpha}$. The held-out target's transfer performances are not used to train the ranker or calibrate the cutoff. For training-free rankers, the ranking rule is fixed by the distance definition, and calibration only estimates the trial budget.

CTC uses conformal prediction to calibrate $\widehat{k}_{\alpha}$ and inherits the corresponding marginal guarantee under exchangeability. Recent work has applied conformal methods to NLP, ranked retrieval and full-ranking uncertainty \citep{xu2025twostage,fermanian2025transductiveconformalinferenceranking, intrator2024streamlining, campos2024conformal}; this metric however uses a conformal procedure as a diagnostic metric for source language rankers.

\paragraph{Implementation.}
Although conformal prediction is valid at any calibration size, the quantile can be too coarse to be operationally useful when $n_{\mathrm{cal}}$ is small \citep{vovk2005algorithmic,shafer2008tutorial,angelopoulos2023conformal}. Since we reserve $20\%$ of non-held-out targets for calibration, small task-model settings yield few calibration scores. The high-coverage quantile can then collapse to the largest observed score, or effectively to trying nearly all candidate sources. This remains statistically valid, but no longer gives an informative budget. We therefore report CTC only for task-model settings with at least $100$ target language queries.

\paragraph{Analysis.}
Table \ref{tab:appendix-cnotc-diagnostics}, read alongside Table \ref{tab:results-pl}, shows that PL and CTC capture different forms of task inequality. PL measures the quality of the first recommendation, while CTC measures the calibrated trial budget needed to obtain a near-oracle source. For example, on Taxi1500, LightGBM is best on both metrics for both models, indicating that it both selects a strong first source and places near-best sources early in the ranking. By contrast, on SIB-200 with XLM-R, genetic distance has the lowest PL but a very large CTC, while MLP shows the opposite trend. This means that genetic distance is better as a one-shot selector on average, but its near-best sources are not reliably concentrated near the top across target languages. On the other hand, MLP's first choice is less strong on average, but it usually brings at least one near-best source into a much shorter shortlist.

To provide a diagnostic of whether the marginal coverage guarantee holds in practice, we compute the empirical percentage of held-out target language folds for which the top-$\widehat{k}_{\alpha}$ shortlist contains a source within $5\%$ relative performance loss of the oracle-best source (denoted, HR). Since we set $\alpha=0.1$, the nominal target is $90\%$. We meet this target for every task using mT5, up to standard error. However, entries using the XLM-R model fall below $90\%$, even when their CTC values are small. This suggests that the issue is not only shortlist length, but also heterogeneity across target language queries. That is, the calibration targets do not fully predict the difficulty of the held-out targets for XLM-R.

\section{Resource Inequality on Unnormalised Performance Loss}
\label{app:unnorm}

Performance loss normalises by oracle performance, which is generally lower on low-resource targets. For example, with mT5, F1 drops from $90.0$ (HRLs) to $71.0$ (LRLs) on SIB-200 and from $69.6$ (HRLs) to $24.3$ (LRLs) on Taxi1500. For completeness, we also measure unnormalised loss and fit a linear mixed-effects model with fixed effects for paradigm, target resource level, their interaction, and model, alongside a random intercept per task/model setting to account for differences in metrics between tasks and models, as well as a variance component for the target languages within each task/model setting. Table~\ref{tab:unnorm-gap} reports model-adjusted LRL-HRL gaps as measured in unnormalised performance loss.

\begin{table}[t]
\centering
\footnotesize
\begin{tabular}{lcc}
\toprule
Paradigm & LRL-HRL gap & $p$ \\
\midrule
English-always & $+4.8$ & $<0.001$ \\
Individual & $+0.8$ & $>0.05$ \\
Composite & $+1.4$ & $>0.05$ \\
Trained & $-4.8$ & $<0.001$ \\
\bottomrule
\end{tabular}
\caption{Model-adjusted LRL-HRL gap in unnormalised performance loss (in task metric points) by paradigm, from a linear mixed-effects model. Positive values indicate higher loss on LRL targets.}
\label{tab:unnorm-gap}
\end{table}

We find that English-always is the only paradigm where unnormalised loss rises significantly on LRL targets. The gaps for individual and composite distances are not significant, and contrarily trained rankers significantly reduce loss on LRL targets.

\section{Testing and Uncertainty Quantification}
\label{app:significance}
To provide statistical background for our claims, we present 95\% confidence intervals (CIs) for our performance losses, as well as testing between different ranker paradigms (trained/composite/individual distance/always-English) for selected results. All CIs and $p$-values are computed by bootstrap, and $p$-values are two-sided. 

In Table \ref{tab:best-vs-english}, we report how the lowest-loss ranker fares against an always-English baseline (for tasks in which it is available), with appropriate uncertainty quantification. This is paired with a more focused comparison between Composite-Equal and LightGBM specifically, against always-English in Table \ref{tab:rankers-vs-english}. One observation worth noting is that in almost all tasks, we see that the lowest-loss ranker is significantly better than the always-English baseline. In other words, choosing a good ranker rather than defaulting to English usually yields a meaningful reduction in transfer loss. 

In terms of the CIs, it is generally the case that the larger the task, the narrower the CI, for both the lowest-loss ranker, as well as always-English. For example, Taxi1500 has substantially more target languages than TyDiQA, and this is reflected in how stable the performance loss is. With that said, more target languages do not seem to have a noticeable effect on making the lowest-loss ranker CI systematically narrower than the always-English one. 

\begin{table*}[!htpb]
\centering
\setlength{\tabcolsep}{3.5pt}
\renewcommand{\arraystretch}{1.06}
\scriptsize

\begin{tabular}{lllcccccc}
\toprule
Task & Model & Lowest-loss ranker
& Ranker PL
& Ranker raw PL
& English PL
& English raw PL
& $\Delta$PL
& $p$ \\
\midrule
\multicolumn{9}{l}{\textit{Tasks with an English baseline}} \\

SIB-200
& mT5
& Script
& 13.0 [10.8, 15.4]
& 8.9 [7.4, 10.5]
& 19.9 [18.1, 21.8]
& 13.9 [12.9, 15.0]
& $-6.9$ [$-9.1$, $-4.5$]
& $<0.001$ \\

& XLM-R
& Genetic
& 23.9 [19.8, 28.2]
& 12.2 [10.2, 14.3]
& 13.0 [11.3, 14.7]
& 6.7 [6.1, 7.2]
& $10.9$ [7.6, 14.5]
& $<0.001$ \\

\midrule

Taxi1500
& mT5
& LightGBM
& 18.9 [17.6, 20.2]
& 5.3 [4.8, 5.9]
& 57.2 [55.8, 58.6]
& 13.2 [12.8, 13.5]
& $-38.3$ [$-40.3$, $-36.3$]
& $<0.001$ \\

& XLM-R
& LightGBM
& 10.0 [8.8, 11.1]
& 2.6 [2.2, 2.9]
& 44.9 [43.6, 46.2]
& 10.0 [9.7, 10.3]
& $-34.9$ [$-36.7$, $-33.1$]
& $<0.001$ \\

\midrule

UD-POS
& mT5
& LightGBM
& 25.6 [21.9, 29.5]
& 8.0 [6.7, 9.3]
& 32.8 [30.0, 35.6]
& 9.9 [9.0, 10.8]
& $-7.2$ [$-11.2$, $-3.2$]
& $<0.001$ \\

& XLM-R
& Composite-RRF
& 15.0 [12.5, 17.8]
& 6.3 [5.1, 7.6]
& 24.8 [21.8, 27.9]
& 9.3 [8.3, 10.3]
& $-9.8$ [$-13.6$, $-6.0$]
& $<0.001$ \\

\midrule

WikiANN
& mT5
& Typological
& 8.9 [7.3, 10.7]
& 6.5 [5.4, 7.7]
& 11.5 [10.1, 12.9]
& 8.6 [7.6, 9.6]
& $-2.6$ [$-4.3$, $-0.9$]
& 0.005 \\

& XLM-R
& Composite-Equal
& 13.4 [10.8, 16.3]
& 8.0 [6.7, 9.4]
& 21.8 [19.0, 24.8]
& 13.5 [12.0, 15.0]
& $-8.4$ [$-11.7$, $-5.1$]
& $<0.001$ \\

\midrule

UD-Dep
& XLM-R
& LightGBM
& 23.8 [19.5, 28.6]
& 10.0 [7.8, 12.4]
& 35.8 [31.4, 40.3]
& 14.1 [12.0, 16.3]
& $-11.9$ [$-16.3$, $-7.5$]
& $<0.001$ \\

\midrule

TyDiQA
& mT5
& Composite-RRF
& 8.6 [3.6, 14.0]
& 5.5 [2.3, 8.8]
& 13.9 [3.6, 26.2]
& 8.8 [2.3, 16.6]
& $-5.3$ [$-13.1$, 1.5]
& 0.152 \\

& XLM-R
& Composite-RRF
& 6.1 [1.3, 11.9]
& 3.5 [0.8, 6.8]
& 16.3 [9.2, 23.9]
& 9.6 [5.7, 13.5]
& $-10.3$ [$-16.6$, $-4.6$]
& $<0.001$ \\

\midrule

XNLI
& mT5
& Genetic
& 3.2 [1.1, 6.9]
& 2.3 [0.8, 4.8]
& 1.5 [0.8, 2.1]
& 1.0 [0.6, 1.5]
& 1.8 [$-0.4$, 5.4]
& 0.257 \\

& XLM-R
& Composite-Equal
& 1.3 [0.5, 2.4]
& 1.0 [0.4, 1.7]
& 3.0 [2.0, 3.9]
& 2.2 [1.5, 2.9]
& $-1.6$ [$-2.6$, $-0.6$]
& 0.003 \\

\midrule

XQuAD
& mT5
& LightGBM
& 21.1 [10.1, 33.7]
& 6.5 [3.1, 10.4]
& 20.9 [13.1, 28.9]
& 6.4 [4.0, 8.8]
& 0.1 [$-10.1$, 11.8]
& 0.986 \\

& XLM-R
& LightGBM
& 32.3 [17.8, 46.5]
& 10.7 [5.7, 15.7]
& 52.5 [44.2, 58.2]
& 18.2 [14.7, 20.9]
& $-20.2$ [$-36.0$, $-5.2$]
& 0.006 \\

\bottomrule
\end{tabular}

\vspace{5pt}

\begin{tabular}{lllcc}
\toprule
Task & Model & Lowest-loss ranker
& Ranker PL
& Ranker raw PL \\
\midrule
\multicolumn{5}{l}{\textit{Tasks without an English baseline}} \\

MasakhaNER
& mT5
& Genetic
& 5.1 [3.2, 7.2]
& 4.0 [2.5, 5.7] \\

& XLM-R
& LightGBM
& 7.9 [4.7, 11.6]
& 3.8 [2.3, 5.5] \\

\midrule

OPUS-100
& mT5
& MLP
& 18.5 [13.4, 23.7]
& 2.1 [1.4, 2.9] \\

\bottomrule
\end{tabular}

\caption{Performance loss for the lowest-loss source-selection ranker in
each task-model setting, with 95\% confidence intervals. PL is relative
performance loss and raw PL is the corresponding unnormalised
oracle-selected-source performance difference. For tasks where English is
available as a source, we additionally report the corresponding English
baseline. $\Delta$PL is the paired difference in relative performance loss,
defined as ranker PL minus English PL, so negative values favour the selected
ranker. The final column reports the paired-test $p$-value. Lower performance
loss is better.}
\label{tab:best-vs-english}
\end{table*}

\begin{table*}[!htpb]
\centering
\setlength{\tabcolsep}{5pt}
\renewcommand{\arraystretch}{1.08}
\scriptsize

\begin{tabular}{lcccc}
\toprule
& \multicolumn{2}{c}{mT5}
& \multicolumn{2}{c}{XLM-R} \\
\cmidrule(lr){2-3}
\cmidrule(lr){4-5}
Task
& Composite-Equal
& LightGBM
& Composite-Equal
& LightGBM \\
\midrule

SIB-200
& \(-4.0^{*}\;[-6.9,-1.0]\)
& \(-3.2^{*}\;[-5.7,-0.6]\)
& \(+14.9^{***}\;[+11.0,+19.0]\)
& \(+11.9^{***}\;[+7.4,+16.7]\) \\

Taxi1500
& \(-22.5^{***}\;[-24.6,-20.5]\)
& \(-38.3^{***}\;[-40.3,-36.3]\)
& \(-15.0^{***}\;[-17.1,-13.0]\)
& \(-34.9^{***}\;[-36.7,-33.1]\) \\

TyDiQA
& \(-3.4\;[-11.9,+3.8]\)
& \(-3.1\;[-10.3,+3.7]\)
& \(-5.3\;[-14.4,+5.2]\)
& \(+0.5\;[-11.4,+14.8]\) \\

UD-Dep
& --
& --
& \(-8.7^{***}\;[-13.3,-3.9]\)
& \(-11.9^{***}\;[-16.3,-7.6]\) \\

UD-POS
& \(-2.4\;[-6.6,+1.7]\)
& \(-7.2^{***}\;[-11.1,-3.2]\)
& \(-9.1^{***}\;[-13.1,-5.0]\)
& \(-5.8^{**}\;[-9.6,-2.0]\) \\

WikiANN
& \(-1.8\;[-3.8,+0.3]\)
& \(-1.7\;[-3.9,+0.6]\)
& \(-8.4^{***}\;[-11.7,-5.2]\)
& \(-7.3^{***}\;[-10.3,-4.4]\) \\

XNLI
& \(+2.4\;[0.0,+6.1]\)
& \(+2.1\;[-1.1,+7.2]\)
& \(-1.6^{**}\;[-2.6,-0.6]\)
& \(-1.6^{**}\;[-2.4,-0.6]\) \\

XQuAD
& \(+19.9^{***}\;[+6.7,+34.8]\)
& \(+0.1\;[-10.1,+11.9]\)
& \(+3.1\;[-0.8,+6.4]\)
& \(-20.2^{**}\;[-36.0,-5.2]\) \\

\bottomrule
\end{tabular}

\caption{Paired differences in performance loss relative to the
English-always baseline, with 95\% confidence intervals. Each entry is
ranker PL minus English PL, so negative values indicate lower performance
loss than transferring always from English.
\(^{*}p\leq .05\), \(^{**}p\leq .01\), \(^{***}p\leq .001\).}
\label{tab:rankers-vs-english}
\end{table*}

Beyond a comparison against always-English, we also present between-paradigm results for each model and task. For this analysis, we average performance loss equally across all rankers within each paradigm for each target language, and compute paired differences between these target-level paradigm means. The results are in Table \ref{tab:pairwise-paradigm}. Both composite and trained methods significantly outperform individual distances in 12 settings, with no significant result in the opposite direction. The comparison between trained and composite rankers is less uniform. Trained rankers significantly outperform composites in five settings, composites significantly outperform trained rankers in one setting, and the paradigms are statistically indistinguishable in 12 settings. This is in line with the results that we see in Table \ref{tab:results-pl}. 

\begin{table*}[!htpb]
\centering
\setlength{\tabcolsep}{9pt}
\renewcommand{\arraystretch}{1.08}
\footnotesize

\begin{tabular}{lcc}
\toprule
Task & mT5 & XLM-R \\
\midrule

\multicolumn{3}{l}{\textit{Composite $-$ Individual}} \\

MasakhaNER
& \(+2.3\;[-1.2,+7.4]\)
& \(-2.1\;[-7.8,+4.7]\) \\

OPUS-100
& \(-9.6^{***}\;[-13.9,-5.1]\)
& -- \\

SIB-200
& \(-0.5\;[-2.7,+1.8]\)
& \(-0.4\;[-3.3,+2.6]\) \\

Taxi1500
& \(-18.1^{***}\;[-19.6,-16.6]\)
& \(-22.4^{***}\;[-24.1,-20.7]\) \\

TyDiQA
& \(-5.2^{***}\;[-8.1,-2.2]\)
& \(-7.6^{***}\;[-14.9,-2.6]\) \\

UD-Dep
& --
& \(-8.2^{***}\;[-11.0,-5.4]\) \\

UD-POS
& \(-9.7^{***}\;[-12.7,-6.8]\)
& \(-9.5^{***}\;[-11.7,-7.3]\) \\

WikiANN
& \(-1.3^{**}\;[-2.3,-0.3]\)
& \(-3.1^{***}\;[-4.4,-1.7]\) \\

XNLI
& \(-2.5^{*}\;[-5.8,-0.1]\)
& \(-0.8^{**}\;[-1.5,-0.2]\) \\

XQuAD
& \(-1.3\;[-11.8,+11.0]\)
& \(-0.2\;[-3.8,+3.3]\) \\

\midrule
\multicolumn{3}{l}{\textit{Trained $-$ Individual}} \\

MasakhaNER
& \(-0.9\;[-3.7,+2.1]\)
& \(-10.5^{***}\;[-15.3,-5.6]\) \\

OPUS-100
& \(-16.8^{***}\;[-20.3,-13.3]\)
& -- \\

SIB-200
& \(-0.4\;[-2.0,+1.2]\)
& \(+0.8\;[-2.8,+4.6]\) \\

Taxi1500
& \(-26.4^{***}\;[-27.9,-24.8]\)
& \(-43.3^{***}\;[-45.0,-41.7]\) \\

TyDiQA
& \(-1.7\;[-7.1,+3.3]\)
& \(-3.8\;[-8.5,+1.4]\) \\

UD-Dep
& --
& \(-9.9^{***}\;[-12.4,-7.4]\) \\

UD-POS
& \(-11.7^{***}\;[-14.6,-8.7]\)
& \(-6.6^{***}\;[-9.1,-4.2]\) \\

WikiANN
& \(-1.5^{**}\;[-2.6,-0.4]\)
& \(-3.5^{***}\;[-4.9,-2.1]\) \\

XNLI
& \(+0.8\;[-2.7,+5.1]\)
& \(-0.5^{*}\;[-1.0,-0.1]\) \\

XQuAD
& \(-14.4^{*}\;[-27.6,-0.8]\)
& \(-23.0^{***}\;[-37.9,-9.3]\) \\

\midrule
\multicolumn{3}{l}{\textit{Trained $-$ Composite}} \\

MasakhaNER
& \(-3.1\;[-9.7,+1.7]\)
& \(-8.4^{*}\;[-18.1,-0.7]\) \\

OPUS-100
& \(-7.3^{***}\;[-11.1,-3.7]\)
& -- \\

SIB-200
& \(+0.1\;[-2.4,+2.6]\)
& \(+1.2\;[-3.1,+5.5]\) \\

Taxi1500
& \(-8.2^{***}\;[-9.9,-6.6]\)
& \(-20.9^{***}\;[-22.8,-19.1]\) \\

TyDiQA
& \(+3.5\;[-2.3,+9.6]\)
& \(+3.8\;[-2.8,+11.3]\) \\

UD-Dep
& --
& \(-1.7\;[-4.1,+0.6]\) \\

UD-POS
& \(-2.0\;[-4.9,+1.1]\)
& \(+2.8^{**}\;[+0.7,+5.2]\) \\

WikiANN
& \(-0.3\;[-1.5,+0.9]\)
& \(-0.4\;[-1.5,+0.8]\) \\

XNLI
& \(+3.3\;[-1.1,+8.0]\)
& \(+0.3\;[-0.2,+0.9]\) \\

XQuAD
& \(-13.1\;[-31.5,+3.2]\)
& \(-22.8^{***}\;[-38.0,-8.3]\) \\

\bottomrule
\end{tabular}

\caption{Pairwise differences in performance loss between source-selection
paradigms, with 95\% confidence intervals.
Each comparison is defined as the first paradigm minus the second, so
negative values indicate lower average performance loss for the first
paradigm.}
\label{tab:pairwise-paradigm}
\end{table*}

\begin{table*}[!htpb]
\centering
\setlength{\tabcolsep}{6pt}
\renewcommand{\arraystretch}{1.08}
\footnotesize

\begin{tabular}{lcccc}
\toprule
& \multicolumn{2}{c}{XLM-R}
& \multicolumn{2}{c}{mT5} \\
\cmidrule(lr){2-3}
\cmidrule(lr){4-5}
Task
& LightGBM & MLP
& LightGBM & MLP \\
\midrule

MasakhaNER
& \textbf{10.7 [5.5, 16.6]}
& 11.1 [5.5, 17.3]
& \textbf{7.1 [3.1, 13.3]}
& 8.5 [5.1, 12.3] \\

OPUS-100
& --
& --
& \textbf{25.0 [19.3, 31.0]}
& 27.1 [21.1, 33.2] \\

SIB-200
& 26.5 [22.0, 31.1]
& \textbf{25.7 [21.3, 30.2]}
& 16.2 [13.5, 19.3]
& \textbf{15.4 [12.8, 18.3]} \\

Taxi1500
& 49.6 [47.7, 51.4]
& \textbf{38.4 [36.5, 40.4]}
& 51.7 [50.1, 53.3]
& \textbf{43.4 [41.7, 45.1]} \\

TyDiQA
& \textbf{11.5 [3.6, 21.5]}
& 16.0 [5.7, 27.5]
& 16.9 [7.4, 28.1]
& \textbf{10.0 [4.9, 15.2]} \\

UD-Dep
& \textbf{22.7 [18.3, 27.3]}
& 26.3 [21.6, 31.1]
& --
& -- \\

UD-POS
& \textbf{11.4 [9.2, 13.8]}
& 16.8 [13.9, 20.0]
& 31.7 [28.2, 35.3]
& \textbf{29.3 [25.8, 32.9]} \\

WikiANN
& \textbf{13.3 [10.5, 16.3]}
& 13.3 [10.6, 16.3]
& \textbf{9.8 [7.9, 11.9]}
& 10.2 [8.1, 12.6] \\

XNLI
& \textbf{1.4 [0.6, 2.4]}
& 1.8 [0.8, 3.1]
& 7.2 [1.9, 14.6]
& \textbf{6.7 [1.6, 14.2]} \\

XQuAD
& \textbf{50.0 [37.9, 59.1]}
& 54.4 [46.6, 60.2]
& 42.5 [25.9, 60.7]
& \textbf{35.3 [22.3, 50.6]} \\

\bottomrule
\end{tabular}

\caption{Mean performance loss with 95\% confidence intervals for trained
rankers under cross-task generalisation. Lower is better. For each held-out
task, rankers are trained on the remaining tasks, with two tasks reserved for
validation. The better trained ranker within each task-model setting is shown
in bold.}
\label{tab:cross-task}
\end{table*}

\section{Cross-Task Generalisation}
In practice, researchers may be interested in seeing how well rankers generalise on an unseen task. To that end, we extend our analysis to mimic this setting for trained rankers. In particular, for each model, we held out one task for testing, randomly selected two of the remaining tasks for validation, and used the remaining tasks for training. One can find the results in Table \ref{tab:cross-task}. 

When we compare against the results for training-free rankers in Table \ref{tab:results-pl} on model/task pairs, we see that under cross-task generalisation, trained rankers attain the lowest PL among the three distance-based paradigms in four of the 18 task-model settings, composite distances attain the lowest PL in six, and individual distances in eight. The trained rankers therefore remain competitive but do not retain their within-task advantage consistently on unseen tasks. This is in line with our expectations, as trained rankers are designed to be trained on one specific task only, and are strongest when transfer observations for a specific task are available. For a previously unseen task, the training-free individual and composite methods are more consistent in these results.

When comparing the trained ranker results when evaluated \emph{in-distribution} (Table \ref{tab:results-pl}) against \emph{out-of-distribution} (Table \ref{tab:cross-task}), we also see that our cross-task setup generally degrades performance. Using the better of LightGBM and MLP in each setting, cross-task generalisation is worse in 13 of 18 task-model settings, better in 4, and essentially unchanged in 1. This is not a universal decrease in performance, but it aligns with our expectation that trained rankers are not robust to distribution/task-shift. If one expects that to be the case, a training-free ranker is likely to be more stable in performance. 

\begin{table}[t]
\centering
\setlength{\tabcolsep}{3pt}
\footnotesize
\begin{tabular}{lcccccc}
\toprule
& \multicolumn{4}{c}{\% dropped} & \multicolumn{2}{c}{LightGBM PL} \\
\cmidrule(lr){2-5}\cmidrule(lr){6-7}
Task & HRL & MRL & LRL & All & Unfilt. & Filt. \\
\midrule
SIB-200 & 14.3 & 4.7 & 7.5 & 7.1 & 18.8 & 19.5 \\
Taxi1500 & 0.0 & 3.4 & 5.3 & 5.2 & 18.9 & 18.4 \\
XNLI & 0.0 & 0.0 & 0.0 & 0.0 & 1.8 & 1.8 \\
UD-POS & 0.0 & 4.8 & 20.0 & 16.0 & 23.8 & 27.7 \\
UD-Dep & 0.0 & 4.8 & 20.0 & 16.0 & -- & -- \\
WikiANN & 0.0 & 4.4 & 20.3 & 15.3 & 11.9 & 11.0 \\
MasakhaNER & -- & -- & 5.0 & 5.0 & 8.0 & 10.4 \\
TyDiQA & 0.0 & 0.0 & 0.0 & 0.0 & 10.1 & 10.1 \\
XQuAD & 0.0 & 0.0 & -- & 0.0 & -- & -- \\
OPUS-100 & 0.0 & 4.8 & 19.6 & 12.1 & 19.4 & 19.8 \\
\bottomrule
\end{tabular}
\caption{\textbf{Left:} proportion of languages dropped from our analysis due to at least one of the six distances being unavailable, by resource level. \textbf{Right:} the performance loss of LightGBM on mT5 LRL targets without (Unfilt.) and with (Filt.) this filtering.}
\label{tab:filtering}
\end{table}

\begin{table}[t]
\centering
\setlength{\tabcolsep}{5pt}
\footnotesize
\begin{tabular}{lccc}
\toprule
Task & Oracle & Random & LightGBM \\
\midrule
SIB-200 & 71.0 & 55.9 & 58.1 \\
Taxi1500 & 24.3 & 12.1 & 19.6 \\
XNLI & 66.9 & 64.1 & 65.7 \\
UD-POS & 29.3 & 15.7 & 22.1 \\
WikiANN & 75.8 & 64.8 & 67.9 \\
MasakhaNER & 79.1 & 70.1 & 70.8 \\
TyDiQA & 63.5 & 53.1 & 57.3 \\
OPUS-100 & 10.6 & 5.7 & 8.8 \\
\bottomrule
\end{tabular}

\vspace{6pt}

\begin{tabular}{llccc}
\toprule
Model & Level & Oracle & SD & Oracle-median gap\\
\midrule
XLM-R & HRL & 70.5 & 13.6 & 17.6 \\
 & MRL & 73.2 & 14.2 & 17.2 \\
 & LRL & 58.7 & 9.6 & 19.2 \\
\midrule
mT5 & HRL & 64.5 & 11.4 & 16.3 \\
 & MRL & 63.7 & 11.4 & 16.0 \\
 & LRL & 57.1 & 8.2 & 12.8 \\
\bottomrule
\end{tabular}
\caption{\textbf{Top:} mean mT5 performance on LRL targets when transferring from the oracle source, a random source (the mean over all candidate sources), and the source selected by LightGBM. XQuAD has no LRL targets. \textbf{Bottom:} mean oracle performance, standard deviation (SD) of performance across candidate sources, and the gap between the oracle and the median source, by model and target resource level.}
\label{tab:lrl-selection}
\end{table}

\section{Language Filtering}
\label{app:filtering}
To ensure every ranker receives the same inputs, our study drops languages for which any of the six distances is unavailable. Table~\ref{tab:filtering} reports the proportion of languages dropped. While we acknowledge that filtering affects LRLs disproportionately, no more than 20.3\% of LRLs, or 16\% of all languages, are ever removed in any task.

As LightGBM handles missing features natively, we evaluate its performance with and without filtering, in order to determine whether this filtering affects our results. We find that LRL performance loss on mT5 changes by at most 3.9 p.p.\ (UD-POS) between the two setups; this consistency therefore suggests that filtering does not impact our resource-level findings.

\section{Source Selection for Low-Resource Targets}
\label{app:lrl-selection}
In theory, a low performance loss on LRL targets could reflect little to potentially gain from source language selection, if hypothetically all candidate sources performed similarly. However, Table~\ref{tab:lrl-selection} demonstrates that this is not true for the tasks we study. In LRL targets on mT5, choosing a source language at random yields performance that falls substantially short of the oracle on most tasks (e.g.\ 12.1 vs.\ 24.3 on Taxi1500). In comparison, LightGBM recovers part of this gap on every task. Although the performance of candidate sources does indeed vary less on LRL targets, the gap between the oracle and the median source remains high at 19.2 (XLM-R) and 12.8 (mT5) points. Therefore, transfer language selection remains a meaningful task even for LRL targets.

\section{Hyperparameters}
\label{app:hyperparams}

\subsection{Model evaluation}

The hyperparameters used for training XLM-R and mT5 across all tasks can be found in Table \ref{tab:hyperparams_xlmr} and \ref{tab:hyperparams_mt5} below.

\begin{table*}[t]
\centering
\scriptsize
\setlength{\tabcolsep}{4pt}
\renewcommand{\arraystretch}{1.05}
\resizebox{\textwidth}{!}{%
\begin{tabular}{lcccccp{4cm}}
\toprule
Dataset & LR & Batch & Limit & Eval Steps & Patience & Notes \\
\midrule
\multicolumn{7}{l}{\textit{Sentence-level classification}} \\
Taxi1500   & \(1{\times}10^{-5}\) & 16 & 10 epochs & 20   & 5  & \\
XNLI       & \(2{\times}10^{-5}\) & 32 & 3 epochs  & 2000 & 5  & \\
\midrule
\multicolumn{7}{l}{\textit{Sequence labeling}} \\
UD POS     & \(5{\times}10^{-5}\) & 16 & 10000 samples        & --   & -- & \\
UD Dep     & \(2{\times}10^{-5}\) (enc.) & 16 & 10000 samples & -- & -- & 256 dims (arc MLP) \\
           & \(2{\times}10^{-4}\) (head) &    &    &    &    & 128 dims (label MLP) \\
\midrule
\multicolumn{7}{l}{\textit{Question answering}} \\
XQuAD      & \(3{\times}10^{-5}\) & 32 & 3 epochs  & --   & -- & \\
TyDi QA     & \(3{\times}10^{-5}\) & 32 & 3 epochs  & --   & -- & \\
\bottomrule
\end{tabular}%
}
\caption{Hyperparameter configurations for XLM-R training. LR = learning rate. Batch = batch size. Limit = maximum number of epochs/steps/samples the model is trained on. Patience = the early stopping patience. SIB-200, WikiANN, and MasakhaNER follow \citet{adelani-etal-2024-sib}, \citet{ebrahimi-etal-2025-model}, and \citet{adelani-etal-2022-masakhaner} respectively. In UD, we train the models on 10000 training samples to handle dataset imbalance and limit training as used in \citet{de-vries-etal-2022-make} for both tasks. We add a biaffine head \citep{dozat-manning-2017-deep} for the dependency parsing task.}
\label{tab:hyperparams_xlmr}
\end{table*}

\begin{table*}[t]
\centering
\scriptsize
\setlength{\tabcolsep}{4pt}
\renewcommand{\arraystretch}{1.05}
\resizebox{\textwidth}{!}{%
\begin{tabular}{lcccccccc}
\toprule
Dataset & LR & Seq Len & Tgt Len & Batch & Grad Accum & Limit & Eval Steps & Patience \\
\midrule
\multicolumn{9}{l}{\textit{Sentence-level classification}} \\
Taxi1500   & \(1{\times}10^{-4}\) & 256  & 4   & 16 & -- & 30 epochs  & 50  & 5  \\
SIB        & \(1{\times}10^{-4}\) & 256  & 4   & 16 & -- & 30 epochs  & 20  & 10 \\
XNLI       & \(3{\times}10^{-5}\) & 1024 & 4   & 32 & 8  & 3 epochs   & 200 & 5  \\
\midrule
\multicolumn{9}{l}{\textit{Sequence labeling}} \\
WikiANN    & \(3{\times}10^{-5}\) & 1024 & 32  & 16 & 2  & 5000 steps & 200 & 5 \\
MasakhaNER & \(3{\times}10^{-5}\) & 1024 & 32  & 16 & 2  & 5000 steps & 200 & 5 \\
UD POS     & \(3{\times}10^{-5}\) & 1024 & 32  & 16 & 2  & 5000 steps & 200 & 5 \\
\midrule
\multicolumn{9}{l}{\textit{Question answering}} \\
XQuAD      & \(3{\times}10^{-5}\) & 1024 & 128 & 16 & 2  & 5000 steps & 200 & 5 \\
TyDi QA     & \(3{\times}10^{-5}\) & 1024 & 128 & 16 & 2  & 5000 steps & 200 & 5 \\
\midrule
\multicolumn{9}{l}{\textit{Machine translation}} \\
OPUS-100       & \(5{\times}10^{-4}\) & 128  & 128 & 64 & -- & 3 epochs   & --  & -- \\
\bottomrule
\end{tabular}%
}
\caption{Hyperparameter configurations for mT5 training. LR = learning rate. Batch = batch size. Limit = maximum number of epochs/steps/samples the model is trained on. Patience = early stopping patience.}
\label{tab:hyperparams_mt5}
\end{table*}

\subsection{Composite-RRF} 
RRF \citep{cormack2009reciprocal} has a smoothing hyperparameter $k$. We set $k=60$, following the original RRF paper, where this value was found to be near-optimal and the method was reported to be relatively insensitive to the exact choice of $k$. 

\subsection{Trained rankers}

Both trained rankers are trained on the six distance features described in Section \ref{sec:rankers}, which are min-max normalised.

\paragraph{LightGBM.} The ranker uses a gradient-boosted decision tree with a LambdaRank objective optimised for NDCG@3, following the architecture in \citet{lin-etal-2019-choosing}. We follow the hyperparameter choices of \citet{ng-etal-2026-modality}, namely setting the number of estimators to 100, number of leaves to 16, learning rate to 0.1, minimum child samples to 10, and L2 regularisation coefficient to 0.2; further performing early stopping with a patience of 25 epochs on validation loss.

\paragraph{MLP.} The MLP consists of two hidden layers of sizes 64 and 32, each followed by ReLU activation and dropout ($p = 0.1$), with a single scalar output. The network is trained with the Adam optimiser (learning rate is $1 \times 10^{-3}$) for up to 200 epochs using a listwise ListNet loss \citep{cao2007learning}, which minimises cross-entropy between the softmax of predicted scores and of actual performance, averaged across target languages. We perform early stopping with a patience of 20 epochs on validation loss. 

\section{Computing Infrastructure}

All model training on XLM-RoBERTa Base (270M parameters) \citep{conneau-etal-2020-unsupervised} and mT5-Base (580M parameters) \citep{xue-etal-2021-mt5} were conducted on 1x NVIDIA H100 80GB GPU, with a total compute time of approximately 750 GPU hours across all training runs and experiments. Experiments on rankers (distance calculation, ranker training, and ranker evaluation) were performed on an Apple M1 Pro with an estimated compute time of 6 hours.

\begin{table}[t]
\centering
\begin{tabular}{lrr}
\toprule
Task & \# Sources & \# Targets \\
\midrule
SIB-200       & 69 & 74  \\
Taxi1500      & 30 & 732 \\
XNLI          & 10 & 10  \\
UD-POS        & 38 & 64  \\
UD-Dep        & 38 & 64  \\
WikiANN       & 34 & 56  \\
MasakhaNER    & 10 & 10  \\
TyDiQA        &  8 &  8  \\
XQuAD         &  9 &  9  \\
OPUS-100      & 26 & 46  \\
\bottomrule
\end{tabular}
\caption{Number of unique source and target languages in the NNRank evaluation.}
\label{tab:nnrank-langs}
\end{table}

\section{Language Coverage of NNRank Comparison}
\label{app:nnrank-langs}

The performance comparison between NNRank, Composite-Equal and LightGBM operates on a smaller subset of languages, where a corresponding language (with an exact match in ISO 639 code) can be found in the eBible corpus \citep{ebible}. Table \ref{tab:nnrank-langs} lists the number of source and target languages represented in this subset.

\section{Artifacts}

The artifacts employed in this study, along with
their respective licenses, are listed in Table \ref{tab:artifacts}.
All artifacts and datasets were used for the purpose of studying language representations and cross lingual transfer, and were handled in accordance with their respective licenses.


\begin{table*}[h]
\centering
\begin{tabular}{ll}
\toprule
\textbf{Artifact} & \textbf{License} \\
\midrule
\multicolumn{2}{l}{\textit{Packages}} \\
URIEL+ \citep{khan2025uriel+, shipton-etal-2026-simple} & CC BY-SA 4.0 \\
Transformers (v5.8.1) \citep{wolf-etal-2020-transformers} & Apache 2.0 \\
PyTorch (v2.10.0) \citep{paszke-2019-torch} & BSD 3-Clause \\
Datasets (v4.8.4) \citep{lhoest-2021-datasets} & Apache 2.0 \\
Evaluate (v0.4.6) \citep{vonwerra-2022-evaluate} & Apache 2.0 \\
LightGBM (v4.6.0) \citep{ke2017lightgbm} & MIT \\
DistaLs \citep{goot-etal-2025-distals} & CC BY-SA 4.0 \\
urielplus-modality-matters \citep{ng-etal-2026-modality} & CC BY-SA 4.0 \\
\midrule
\multicolumn{2}{l}{\textit{Datasets}} \\
Taxi1500 \citep{ma-etal-2025-taxi1500} & Apache 2.0 \\
SIB200 \citep{adelani-etal-2024-sib} & CC BY-SA 4.0 \\
XNLI \citep{conneau-etal-2018-xnli} & CC BY-NC 4.0 \\
WikiANN \citep{pan-etal-2017-cross} & Apache 2.0 \\
MasakhaNER 2.0 \citep{adelani-etal-2022-masakhaner} & CC BY-NC 4.0 \\
Universal Dependencies (v2.14) \citep{ud} & Various \\
OPUS-100 \citep{zhang-etal-2020-improving} & Various \\
XQuAD \citep{Artetxe:etal:2019} & CC BY-SA 4.0 \\
TyDi QA \citep{clark-etal-2020-tydi} & Apache 2.0  \\
\midrule
\multicolumn{2}{l}{\textit{Models}} \\
XLM-RoBERTa Base \citep{conneau-etal-2020-unsupervised} & MIT \\
mT5 Base \citep{xue-etal-2021-mt5} & Apache 2.0 \\
\bottomrule
\end{tabular}
\caption{Artifacts used in this study, and their licenses.}
\label{tab:artifacts}
\end{table*}

\section{Usage of AI Assistants}

We used generative AI only in a limited capacity. Namely, generating auto-code completions, which were manually verified. They were also used for proofreading texts.

\end{document}